\documentclass{article} %
\usepackage[table, dvipsnames]{xcolor}
\usepackage[preprint]{colm2025_conference}
\usepackage{microtype}
\usepackage{hyperref}
\usepackage{url}
\usepackage{tabularx}
\usepackage{booktabs}       %
\usepackage{amsthm,amsmath,amssymb,amsfonts}
\usepackage{lineno}

\usepackage{amsthm,amsmath,amssymb,amsfonts,mathbbol}
\usepackage{wrapfig}
\usepackage{caption}
\usepackage{subcaption}
\usepackage{colortbl}
\usepackage[most]{tcolorbox}
\usepackage{multirow}
\usepackage{arydshln}
\usepackage{stackengine}  %
\usepackage{xspace}
\usepackage{soul} 

\usepackage{algorithm}
\usepackage[noabbrev, capitalise]{cleveref}
\usepackage{mathtools}
\usepackage[export]{adjustbox}

\usepackage{tikz}
\usetikzlibrary{decorations.pathmorphing,shapes}
\usetikzlibrary{arrows,arrows.meta,bending,positioning,patterns,patterns.meta,calc}
\usepackage{scalerel,stackengine}
\usetikzlibrary{circuits.logic.US}
\usetikzlibrary{shapes.misc}
\usetikzlibrary{shapes.arrows}
\usetikzlibrary{arrows,shapes.geometric}
\usetikzlibrary{positioning}
\tikzset{tips=proper,edge/.style = {->,>=latex'},lbl/.style={draw=none,font={\footnotesize\ttfamily}}}
\usepackage{pgfplots}
\usepackage{pgfplotstable}
\pgfplotsset{compat=newest}
\usepgfplotslibrary{fillbetween,groupplots}
\newcommand{\implws}{\textvisiblespace}
\usepackage{pgffor}
\usepackage{ifthen}
\newcounter{sarrow}

\usepackage{multirow}

\definecolor{palette-orange}{HTML}{ff3a20}
\definecolor{palette-green}{HTML}{5b8c5a}
\definecolor{palette-blue}{HTML}{0e79b2}
\definecolor{palette-yellow}{HTML}{f5b700}
\definecolor{palette-green}{HTML}{1e2f23}
\definecolor{palette-purple}{HTML}{331832}
\definecolor{dark gray}{HTML}{808080}
\definecolor{darker gray}{HTML}{606060}

\definecolor{palette-alt-green}{HTML}{084c61}
\definecolor{palette-alt-pink}{HTML}{d11149}
\definecolor{palette-alt-yellow}{HTML}{e3b505}
\definecolor{palette-alt-purple}{HTML}{331832}
\definecolor{palette-alt-blue}{HTML}{01baef}
\definecolor{palette-alt-gray}{HTML}{808f85}

\definecolor{sanae5}{rgb}{0.1750865648952205, 0.11840023306916837, 0.24215989137836502}
\definecolor{sanae1}{rgb}{0.8559578605899612, 0.6418993116910497, 0.6754191211563135}
\definecolor{pastel-red}{HTML}{FF6961}
\definecolor{red-red}{HTML}{DB1F48}
\definecolor{pastel-red}{HTML}{FF6961}
\definecolor{spearmint}{HTML}{4EB6B0}
\pgfplotscreateplotcyclelist{colorpalette}{
{palette-orange,fill=palette-orange},
{palette-green ,fill=palette-green },
{palette-blue  ,fill=palette-blue  },
{palette-purple,fill=palette-purple}}

\pgfplotsset{select coords between index/.style 2 args={
    x filter/.code={
        \ifnum\coordindex<#1\fi
        \ifnum\coordindex>#2\fi
    }
}}

\pgfplotsset{compat=1.17}

\pgfplotstableread[col sep=semicolon,trim cells]{
t ; prob
0 ; 0.03923155305405986
1 ; 0.024577278468636495
2 ; 0.007276373877266687
3 ; 0.028308393891704593
4 ; 0.01580119349294541
5 ; 0.03186393344885731
6 ; 0.0012056173438990561
7 ; 0.019276390394675352
8 ; 0.012433867274512891
9 ; 0.02542855106567576
10 ; 0.016951222565980643
11 ; 0.030429487940590555
12 ; 0.024612396718092032
13 ; 0.02369028640856639
14 ; 0.025795558945475677
15 ; 3.451571497620609e-05
16 ; 0.019473661166120298
17 ; 0.022244166422540554
18 ; 0.02669973533714404
19 ; 0.00953979975083729
20 ; 0.031453533861785295
21 ; 0.025928634532335872
22 ; 0.017656447322222393
23 ; 0.02975749991546848
24 ; 0.019681391534660465
25 ; 0.024485404188457894
26 ; 0.007581783519911938
27 ; 0.01121030713639992
28 ; 0.009406078083906423
29 ; 0.03943046985928758
30 ; 0.025326412143107012
31 ; 0.025835882714199485
32 ; 0.03519289692819652
33 ; 0.02234979472518933
34 ; 0.01867495738873156
35 ; 0.009421041516853192
36 ; 0.009380211636437319
37 ; 0.025390539103420852
38 ; 0.01298919484090052
39 ; 0.016732768293271937
40 ; 0.0014596041850309925
41 ; 0.005002869773278239
42 ; 0.039174768272527735
43 ; 0.023304624240336923
44 ; 0.025774213594779775
45 ; 0.027325497276596163
46 ; 0.007691895231113642
47 ; 0.015872389706587414
48 ; 0.0286616818849944
49 ; 0.0029732233074537063
}\data
\pgfplotstableread[col sep=semicolon,trim cells]{
t ; prob
0 ; 0
1 ; 0.1
2 ; 0.4
3 ; 0.5
4 ; 0.0
5 ; 0.0
}\da
\pgfplotstableread[col sep=semicolon,trim cells]{
t ; prob
0 ; 0
1 ; 0.7
2 ; 0.25
3 ; 0.05
4 ; 0.0
5 ; 0.0
}\daa
\pgfplotstableread[col sep=semicolon,trim cells]{
    t ; prob
    16 ; 0.08580414270114088
    17 ; 0.09801144292826634
    18 ; 0.1176434098040447
    19 ; 0.042033921211756926
    20 ; 0.13858942521949394
    21 ; 0.1142458132797797
    22 ; 0.07779720067569487
    23 ; 0.1311164216833674
    24 ; 0.08671943675055112
    25 ; 0.10788670384882341
}\dat

\pgfplotstableread[col sep=semicolon,trim cells]{
    t ; prob
    22 ; 0.017656447322222393
    23 ; 0.02975749991546848
    24 ; 0.019681391534660465
    25 ; 0.024485404188457894
    26 ; 0.007581783519911938
    27 ; 0.01121030713639992
    28 ; 0.009406078083906423
    29 ; 0.03943046985928758
    30 ; 0.025326412143107012
    31 ; 0.025835882714199485
    32 ; 0.03519289692819652
    33 ; 0.02234979472518933
    34 ; 0.01867495738873156
    35 ; 0.009421041516853192
    36 ; 0.009380211636437319
    37 ; 0.025390539103420852
    38 ; 0.01298919484090052
    39 ; 0.016732768293271937
    40 ; 0.0014596041850309925
}\constraint

\pgfplotstableread[col sep=semicolon,trim cells]{
    t ; prob
    22 ; 0.07779720067569487
    23 ; 0.1311164216833674
    24 ; 0.08671943675055112
    25 ; 0.10788670384882341
}\constraintclipped

\pgfplotstableread[col sep=semicolon, trim cells]{
    x ; y
    20 ; 0.031453533861785295
    20 ; 1.02e-5
    21 ; 1.02e-5
    21 ; 0.031453533861785295
}\wrong

\pgfplotstableread[col sep=semicolon, trim cells]{
    x ; y
    20 ; 0.13858942521949394
    20 ; 1.02e-5
    21 ; 1.02e-5
    21 ; 0.13858942521949394
}\wrongtwo

\pgfplotstableread[col sep=semicolon, trim cells]{
    x ; y
    23 ; 0.13858942521949394
    23 ; 1.02e-5
    24 ; 1.02e-5
    24 ; 0.13858942521949394
}\correct

\pgfmathdeclarefunction{gaussian}{2}{%
  \pgfmathparse{1/(#2*sqrt(2*pi))*exp(-((x-#1)^2)/(2*#2^2))}%
}

\usepackage{algorithmic}

\usepackage{caption}
\usepackage{multicol}

\definecolor{sanae0}{rgb}{0.9312692223325372, 0.8201921796082118, 0.7971480974663592}
\definecolor{sanae1}{rgb}{0.8559578605899612, 0.6418993116910497, 0.6754191211563135}
\definecolor{sanae2}{rgb}{0.739734329496642, 0.4765280683170713, 0.5959617419736206}
\definecolor{sanae3}{rgb}{0.57916573903086, 0.33934576125314425, 0.5219003947563425}
\definecolor{sanae4}{rgb}{0.37894937987024996, 0.2224702044652721, 0.41140014301575434}
\definecolor{sanae5}{rgb}{0.1750865648952205, 0.11840023306916837, 0.24215989137836502}
\definecolor{pastel-red}{HTML}{FF6961}
\definecolor{spearmint}{HTML}{4EB6B0}
\definecolor{teal}{HTML}{12486B}
\definecolor{cyan}{HTML}{9CE8F1}
\definecolor{violett}{HTML}{cdb4db}

\definecolor{darkblue}{rgb}{0, 0, 0.5}
\hypersetup{colorlinks=true, linkcolor=teal,citecolor=Cerulean, urlcolor=teal}

\usepackage{amsmath,amsfonts,bm}

\def\eqref#1{equation~\ref{#1}}

\def\1{\bm{1}}

\def\rs{{\textnormal{s}}}

\def\ry{{\textnormal{y}}}

\def\rvs{{\mathbf{s}}}

\def\rvy{{\mathbf{y}}}
\def\rvY{{\mathbf{Y}}}

\DeclareMathAlphabet{\mathsfit}{\encodingdefault}{\sfdefault}{m}{sl}
\SetMathAlphabet{\mathsfit}{bold}{\encodingdefault}{\sfdefault}{bx}{n}

\def\sV{{\mathbb{V}}}

\DeclareMathOperator*{\argmax}{arg\,max}

\newcommand{\p}{\ensuremath{p}}
\newcommand{\q}{\ensuremath{q}}
\newcommand{\wrt}{w.r.t.\ }
\newcommand{\conditional}[2]{#1_{|#2}}
\newcommand{\wolog}{w.l.o.g}
\newcommand{\ch}{\ensuremath{\mathsf{in}}}
\newcommand{\vars}{\ensuremath{\mathsf{vars}}}

\newcommand{\phiattr}{\phi_{\bm{a}}}
\newcommand{\gradphiattr}{\nabla \phi_{\bm{a}}}
\newcommand{\stilde}{\tilde{\rvs}}
\newcommand{\condmarg}{\tilde{\p}_{\mathtt{cond}}}
\newcommand{\attrib}{\bm{a}}

\newcommand{\continuation}{\rvy_{i+1:T}}
\newcommand{\fully}{\rvy_{1:T}}
\newcommand{\yprefix}{\rvy_{1:i}}

\newcommand{\ytilde}{\Tilde{\rvy}}

\title{Semantic Probabilistic Control of Language Models}

\author{Kareem Ahmed\textsuperscript{*}, Catarina G. Belem\textsuperscript{*}, Padhraic Smyth \& Sameer Singh\\
Department of Computer Science\\
University of California, Irvine\\
\texttt{\{ahmedky, cbelem, smyth, sameer\}@uci.edu} \\
}

\newcommand{\eg}[0]{\textit{e.g.},\xspace}
\newcommand{\ie}[0]{\textit{i.e.},\xspace}

\newcommand{\llama}{\texttt{Llama-3.2 (1B)}\xspace}
\newcommand{\gptimdb}{\texttt{GPT2-IMDB}\xspace}
\newcommand{\perplexitymodel}{\texttt{\texttt{Meta-Llama-3-70B}}\xspace}

\newcommand{\phitoxicity}{$\phi_{\mathtt{toxicity}}$\xspace}

\newcommand{\phisentiment}{$\phi_{\mathtt{sentiment}}$\xspace}

\newcommand{\phitopic}{$\phi_{\mathtt{topic}}$\xspace}

\newcommand{\realtoxicityprompts}{\texttt{RealToxicityPrompts}\xspace}
\newcommand{\imdb}{\texttt{IMDB}\xspace}
\newcommand{\topicdataset}{\texttt{TopicAnnotations-Llama-3.1-405B-FP8}\xspace}
\newcommand{\random}{\texttt{random}\xspace}
\newcommand{\beamsearch}{\texttt{beamsearch}\xspace}
\newcommand{\bestofn}{\texttt{BoN}\xspace}

\newcommand{\ours}{\texttt{{\color{spearmint}S}{\color{palette-blue}Con}{\color{violett}E}}\xspace}
\newcommand{\oursc}{\texttt{{\color{spearmint}S}{\color{palette-blue}Con}{\color{violett}E}}\xspace}

\newif\ifcomments
\commentstrue
\ifcomments
\providecommand{\kat}[2][]{{\protect\color{teal}{[Kat:\textbf{#1} #2]}}}
\providecommand{\kareem}[2][]{{\protect\color{orange}{[Kareem:\textbf{#1} #2]}}}
\providecommand{\sameer}[2][]{{\protect\color{purple}{[Sameer:\textbf{#1} #2]}}}
\providecommand{\padhraic}[2][]{{\protect\color{orange}{[Padhraic:\textbf{#1} #2]}}}
\else
\providecommand{\kat}[2][]{}
\providecommand{\kareem}[2][]{}
\providecommand{\sameer}[2][]{}
\providecommand{\padhraic}[2][]{}
\fi

\begin{document}

\ifcolmsubmission
\linenumbers
\fi
\maketitle
\def\thefootnote{*}\footnotetext{Co-first authors.}\def\thefootnote{\arabic{footnote}}
\begin{abstract}
Semantic control entails steering LM generations towards satisfying subtle non-lexical
constraints\textemdash \eg toxicity, sentiment, or politeness\textemdash attributes that
can be captured by a sequence-level \emph{verifier}.
It can thus be viewed as sampling from the LM distribution conditioned on the target
attribute, a computationally intractable problem due to the non-decomposable nature
of the verifier.
Existing approaches to LM control either only deal with syntactic constraints which
cannot capture the aforementioned attributes, or rely on sampling to explore the
conditional LM distribution, an ineffective estimator for low-probability events.
In this work, we leverage a verifier's gradient information to efficiently reason
over \emph{all} generations that satisfy the target attribute, enabling precise
steering of LM generations by reweighing the next-token distribution.
Starting from an initial sample, we create a local LM distribution favoring semantically
similar sentences.
This approximation enables the tractable computation of an \emph{expected sentence embedding}.
We use this expected embedding, informed by the verifier's evaluation at the initial sample, to estimate the probability of satisfying the constraint, which directly informs the update to the next-token distribution.
We evaluated the effectiveness of our approach in controlling the toxicity, sentiment, and topic-adherence of LMs yielding generations satisfying the constraint with high probability ($>95\%$) without degrading their quality.
\end{abstract}

\section{Introduction}
\label{sec:introduction}
Despite the unprecedented capabilities of language models (LMs), steering their
generations towards specific syntactic or semantic constraints remains an unsolved challenge~\citep{sun2023evaluating, Liu2024Structured}. %
Syntactic (or \emph{lexical}) constraints define at each position in the sequence
the set of admissible tokens that, taken together, constitute a valid string under
the constraint.
A common use case for such constraints is to generate output in some formal language,
for example, structured data, API calls, or code snippets~\citep{geng2025jsonschemabenchrigorousbenchmarkstructured}.
Syntactic constraints are \emph{easy} to deal with in a very precise sense: through knowledge
compilation~\citep{Darwiche2002knowledge}, we can efficiently capture the computation graph of generations satisfying the constraint, which we can then proceed to \emph{probabilistically} reason about, 
exactly when possible~\citep{Ahmed2022Semantic}, 
and otherwise approximately~\citep{outlines, Honghua2024Adaptable, koo2024, guidance, Ahmed2025controllable}.

Semantic (or \textit{non-lexical}) constraints, on the other hand, are often defined in terms of sequence-level, non-decomposable classifiers, or \textit{verifiers}, often complex neural networks, that assign non-negative scores to sequences of tokens.
In that sense, semantic constraints are doubly hard: we have to contend with not only the
hardness of probabilistic reasoning but also the lack of a tractable representation of the
constraint over which to reason.
Semantic constraints encompass use cases in which we might wish to control
sequence-level properties of generations that are hard to capture in formal language,
\eg controlling toxicity, sentiment, or topic in creative writing; targeting outputs deemed 
favorable by a verifier in reasoning tasks, or generating \emph{correct} code that exhibits
stylistic requirements~\citep{geng2025jsonschemabenchrigorousbenchmarkstructured}.

Existing approaches to semantic control of LMs therefore generally fall into one of two classes:
sample-reweigh and sequential Monte Carlo (SMC) approaches, each of which suffers from
major drawbacks.
Sample-reweigh, prominently known as best-of-n~\citep{Stiennon2020-BestOfN-2020-Neurips}, generates complete sequences that are reweighed
by the potential function, returning the only highest scoring sequence.
Sample-reweigh does not factor in the constraint during generation and therefore the number of samples needed to satisfy the constraint can grow exponentially, especially for very low probability constraints.
SMC, on the other hand, maintains a set of samples that evolve through time, factoring in
the likelihood of the new sample under the model as well as information about the constraint at every step of generation, either through learning twist functions~\citep{Zhao-et-al-2024-twist-functions-2024-ICML}
or evaluating the potential function on partial sequences~\citep{loula2025syntactic-2025-ICLR}. 
SMC, however, is not without its own drawbacks: it requires a large number of samples, which can grow exponentially with the dimensionality of the space;
it suffers from sample impoverishment, where after a few iterations only a few samples carry almost all the weight, with resampling, while addressing degeneracy, leading to a loss of sample diversity;
and crucially, it requires the careful design of a proposal distribution, which greatly affects the performance of SMC. 

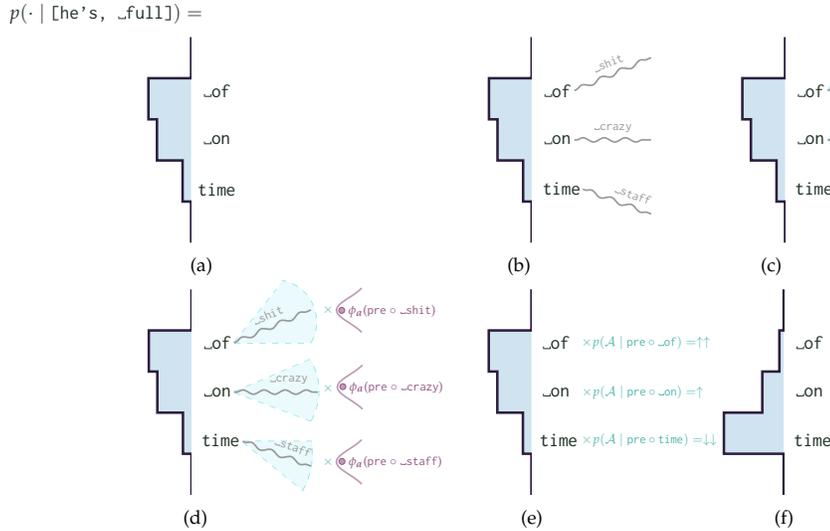
\begin{figure*}[!t]
\centering
\begin{adjustbox}{minipage=\linewidth,scale=0.80}
\begin{subfigure}[t]{0.3\textwidth}
 \centering
     \begin{tikzpicture}
    \hspace{-15mm}%
    \node[palette-green] (canon) at (0.0, 3.8) {\footnotesize$\p( \mathbf{\cdot} \mid\texttt{[he's, \implws full]}) =$ };
    \begin{axis}[
        const plot,
        width=5cm,
        height=3cm,
        ymax=1,ymin=0,
        xmin=0,xmax=5,
        log origin=infty,
        ticks=none,
        axis lines*=left,
        axis line style={draw=none},
        clip mode=individual,
        rotate around={90:(current axis.origin)},
    ]
    \addplot[forget plot,name path=pr, draw=none, very thick, fill=palette-blue!20!white] table [x=t,y=prob,palette-blue] from \da \closedcycle;
    \addplot[forget plot,name path=pr, very thick, sanae5,] table [x=t,y=prob,palette-blue] from \da;
    \node[palette-green] (token1) at (3.7, -0.3) {\footnotesize\texttt{\implws of}};
    \node[palette-green] (token2) at (2.5, -0.3) {\footnotesize\texttt{\implws on}};
    \node[palette-green] (token3) at (1.3, -0.3) {\footnotesize\texttt{time}};
    \end{axis}
\end{tikzpicture}
\caption{}
\end{subfigure}%
\hspace{10mm}
\begin{subfigure}[t]{0.3\textwidth}
 \centering
     \begin{tikzpicture}
    \hspace{-11.2mm}%
    \begin{axis}[
        const plot,
        width=5cm,
        height=3cm,
        ymax=1,ymin=0,
        xmin=0,xmax=5,
        log origin=infty,
        ticks=none,
        axis lines*=left,
        axis line style={draw=none},
        clip mode=individual,
        rotate around={90:(current axis.origin)},
    ]
    \addplot[forget plot,name path=pr, draw=none, very thick, fill=palette-blue!20!white] table [x=t,y=prob,palette-blue] from \da \closedcycle;
    \addplot[forget plot,name path=pr, very thick, sanae5] table [x=t,y=prob,palette-blue] from \da;
    \node[palette-green] (token1) at (3.7, -0.3) {\footnotesize\texttt{\implws of}};
    \node[palette-green] (token2) at (2.5, -0.3) {\footnotesize\texttt{\implws on}};
    \node[palette-green] (token3) at (1.325, -0.35) {\footnotesize\texttt{time}};

    \draw[-, gray!80,decorate,decoration={snake,amplitude=1pt,pre length=1pt,post length=0pt}, thick] (3.7,-0.5) -- (4.5,-1.4) node[midway, above, sloped, rotate=90] (shit) {\tiny\texttt{\implws shit}};
    \draw[-, gray!80,decorate,decoration={snake,amplitude=1pt,pre length=1pt,post length=0pt}, thick] (2.5,-0.5) -- (2.5,-1.4) node[midway, above, sloped, rotate=90] (crazy) {\tiny\texttt{\implws crazy}};
    \draw[-, gray!80,decorate,decoration={snake,amplitude=1pt,pre length=1pt,post length=0pt}, thick] (1.3, -0.6) -- (0.7,-1.4) node[midway, above, sloped, rotate=-90] (employed) {\tiny\texttt{\hspace{2em}\implws staff}};
    \end{axis}
\end{tikzpicture}
\caption{}
\end{subfigure}%
\begin{subfigure}[t]{0.3\textwidth}
 \centering
     \begin{tikzpicture}
     \hspace{-11mm}
    \begin{axis}[
        const plot,
        width=5cm,
        height=3cm,
        ymax=1,ymin=0,
        xmin=0,xmax=5,
        log origin=infty,
        ticks=none,
        axis lines*=left,
        axis line style={draw=none},
        clip mode=individual,
        rotate around={90:(current axis.origin)},
    ]
    \addplot[forget plot,name path=pr, draw=none, very thick, fill=palette-blue!20!white] table [x=t,y=prob,palette-blue] from \da \closedcycle;
    \addplot[forget plot,name path=pr, very thick, sanae5] table [x=t,y=prob,palette-blue] from \da;
    \node[palette-green] (token1) at (3.7, -0.3) {\footnotesize\texttt{\implws of}};
    \node[palette-green] (token2) at (2.5, -0.3) {\footnotesize\texttt{\implws on}};
    \node[palette-green] (token3) at (1.325, -0.34) {\footnotesize\texttt{time}};

    \draw[thick, -, cyan, dashed] (3.7,-0.5) -- (3.7,-1.4);
    \draw[thick, -, cyan, dashed] (3.7,-0.5) -- (5.0,-1.0);
    \fill[cyan!20] (3.7,-0.5) -- (3.7,-1.4) arc (240:350:0.7cm) -- (5.0,-1.0) -- cycle;
    \draw[cyan, dashed] (3.7,-1.4) arc (240:350:0.7cm); %
    \draw[-, gray!80,decorate,decoration={snake,amplitude=1pt,pre length=1pt,post length=0pt}, thick] (3.7,-0.5) -- (4.5,-1.4) node[midway, above, sloped, rotate=90] (shit) {\tiny\texttt{\implws shit}};

    \draw[thick, -, cyan, dashed] (2.5,-0.5) -- (3.3,-1.4);
    \draw[thick, -, cyan, dashed] (2.5,-0.5) -- (1.8,-1.4);
    \fill[cyan!20] (2.5,-0.5) -- (3.3,-1.4) arc (300:240:1.0cm) -- (1.8,-1.4) -- cycle;
    \draw[cyan, dashed] (1.8,-1.4) arc (240:300:1.0cm); %
    \draw[-, gray!80,decorate,decoration={snake,amplitude=1pt,pre length=1pt,post length=0pt}, thick] (2.5,-0.5) -- (2.5,-1.48) node[midway, above, sloped, rotate=90] (crazy) {\tiny\texttt{\hspace{2em}\implws crazy}};

    \draw[thick, -, cyan, dashed] (1.3, -0.6) -- (1.3,-1.41);
    \draw[thick, -, cyan, dashed] (1.3, -0.6) -- (0.1,-1.31);
    \fill[cyan!20] (1.3, -0.6) -- (1.3,-1.41) arc (280:240:1.2cm) -- (0.1,-1.31) -- cycle;
    \draw[cyan, dashed] (0.1,-1.3) arc (240:280:1.22cm);

    \draw[-, gray!80,decorate,decoration={snake,amplitude=1pt,pre length=1pt,post length=0pt}, thick] (1.3, -0.6) -- (0.7,-1.4) node[midway, above, sloped, rotate=-90] (employed) {\tiny\texttt{\hspace{2em}\implws staff}};
    \end{axis}
\end{tikzpicture}
\caption{}
\end{subfigure}%

\hspace{-45pt}
\begin{subfigure}[t]{0.5\textwidth}
 \centering
     \begin{tikzpicture}
    \begin{axis}[
        const plot,
        width=5cm,
        height=3cm,
        ymax=1,ymin=0,
        xmin=0,xmax=5,
        log origin=infty,
        ticks=none,
        axis lines*=left,
        axis line style={draw=none},
        clip mode=individual,
        rotate around={90:(current axis.origin)},
    ]
    \addplot[forget plot,name path=pr, draw=none, very thick, fill=palette-blue!20!white] table [x=t,y=prob,palette-blue] from \da \closedcycle;
    \addplot[forget plot,name path=pr, very thick, sanae5] table [x=t,y=prob,palette-blue] from \da;
    \node[palette-green] (token1) at (3.7, -0.3) {\footnotesize\texttt{\implws of}};
    \node[palette-green] (token2) at (2.5, -0.3) {\footnotesize\texttt{\implws on}};
    \node[palette-green] (token3) at (1.325, -0.34) {\footnotesize\texttt{time}};

    \draw[thick, -, cyan, dashed] (3.7,-0.5) -- (3.7,-1.4);
    \draw[thick, -, cyan, dashed] (3.7,-0.5) -- (5.0,-1.0);
    \fill[cyan!20] (3.7,-0.5) -- (3.7,-1.4) arc (240:350:0.7cm) -- (5.0,-1.0) -- cycle;
    \draw[cyan, dashed] (3.7,-1.4) arc (240:350:0.7cm); %
    \draw[-, gray!80,decorate,decoration={snake,amplitude=1pt,pre length=1pt,post length=0pt}, thick] (3.7,-0.5) -- (4.5,-1.4) node[midway, above, sloped, rotate=90] (shit) {\tiny\texttt{\implws shit}};

    \draw[thick, -, cyan, dashed] (2.5,-0.5) -- (3.3,-1.4);
    \draw[thick, -, cyan, dashed] (2.5,-0.5) -- (1.8,-1.4);
    \fill[cyan!20] (2.5,-0.5) -- (3.3,-1.4) arc (300:240:1.0cm) -- (1.8,-1.4) -- cycle;
    \draw[cyan, dashed] (1.8,-1.4) arc (240:300:1.0cm); %
    \draw[-, gray!80,decorate,decoration={snake,amplitude=1pt,pre length=1pt,post length=0pt}, thick] (2.5,-0.5) -- (2.5,-1.48) node[midway, above, sloped, rotate=90] (crazy) {\tiny\texttt{\hspace{2em}\implws crazy}};

    \draw[thick, -, cyan, dashed] (1.3, -0.6) -- (1.3,-1.41);
    \draw[thick, -, cyan, dashed] (1.3, -0.6) -- (0.1,-1.31);
    \fill[cyan!20] (1.3, -0.6) -- (1.3,-1.41) arc (280:240:1.2cm) -- (0.1,-1.31) -- cycle;
    \draw[cyan, dashed] (0.1,-1.3) arc (240:280:1.22cm);
    \draw[-, gray!80,decorate,decoration={snake,amplitude=1pt,pre length=1pt,post length=0pt}, thick] (1.3, -0.6) -- (0.7,-1.4) node[midway, above, sloped, rotate=-90] (employed) {\tiny\texttt{\hspace{2em}\implws staff}};

    \node[palette-green] (phi1) at (4.475, -2.29) {\color{sanae3}{\tiny\tikz\draw[fill=sanae3!60] (0,0) circle (.5ex); $\phiattr$(\texttt{pre} $\circ$ \texttt{\implws shit})}};
    \node[palette-green] (phi2) at (2.62, -2.34) {\color{sanae3}{\tiny \tikz\draw[fill=sanae3!60] (0,0) circle (.5ex); $\phiattr$(\texttt{pre} $\circ$ \texttt{\implws crazy})}};
    \node[palette-green] (phi3) at (0.8, -2.33) {\color{sanae3}{\tiny \tikz\draw[fill=sanae3!60] (0,0) circle (.5ex); $\phiattr$(\texttt{pre} $\circ$ \texttt{\implws staff})}};

    \node[palette-green] (times1) at (2.6, -1.6) {\tiny\color{spearmint} {$\times$}};
    \node[palette-green] (times1) at (4.5, -1.6) {\tiny\color{spearmint} {$\times$}};
    \node[palette-green] (times1) at (0.8, -1.6) {\tiny\color{spearmint} {$\times$}};
    
    \end{axis}
    \begin{scope}[rotate around={90:(current axis.origin)}, xshift=0.59\linewidth, yshift=-0.405\linewidth] 
    \begin{axis}[
    no markers, domain=0:10, samples=100,
    axis lines*=left,
    every axis y label/.style={at=(current axis.above origin),anchor=south},
    every axis x label/.style={at=(current axis.right of origin),anchor=west},
    height=2cm,
    xtick=\empty, ytick=\empty,
    enlargelimits=false, clip=false, axis on top,
    grid = major,
    axis lines*=left,
    axis line style={draw=none},
    ]
    \addplot [thick,sanae3!60, domain=1.5:8.5] {gaussian(5,2)};
\end{axis}
\end{scope}
\begin{scope}[rotate around={90:(current axis.origin)}, xshift=0.425\linewidth, yshift=0.385\linewidth] 
    \begin{axis}[
    no markers, domain=0:10, samples=100,
    axis lines*=left,
    every axis y label/.style={at=(current axis.above origin),anchor=south},
    every axis x label/.style={at=(current axis.right of origin),anchor=west},
    height=2cm,
    xtick=\empty, ytick=\empty,
    enlargelimits=false, clip=false, axis on top,
    grid = major,
    axis lines*=left,
    axis line style={draw=none},
    ]
    \addplot [thick,sanae3!60, domain=1.5:8.5] {gaussian(5,2)};
\end{axis}
\end{scope}
    \begin{scope}[rotate around={90:(current axis.origin)}, xshift=0.432\linewidth, yshift=0.206\linewidth] 
    \begin{axis}[
    no markers, domain=0:10, samples=100,
    axis lines*=left,
    every axis y label/.style={at=(current axis.above origin),anchor=south},
    every axis x label/.style={at=(current axis.right of origin),anchor=west},
    height=2cm,
    xtick=\empty, ytick=\empty,
    enlargelimits=false, clip=false, axis on top,
    grid = major,
    axis lines*=left,
    axis line style={draw=none},
    ]
    \addplot [thick,sanae3!60, domain=1.5:8.5] {gaussian(5,2)};
\end{axis}
\end{scope}
\end{tikzpicture}
\caption{}
\end{subfigure}%
\begin{subfigure}[t]{0.3\textwidth}
 \centering
     \begin{tikzpicture}
     \hspace{-38pt}
    \begin{axis}[
        const plot,
        width=5cm,
        height=3cm,
        ymax=1,ymin=0,
        xmin=0,xmax=5,
        log origin=infty,
        ticks=none,
        axis lines*=left,
        axis line style={draw=none},
        clip mode=individual,
        rotate around={90:(current axis.origin)},
    ]
    \addplot[forget plot,name path=pr, draw=none, very thick, fill=palette-blue!20!white] table [x=t,y=prob,palette-blue] from \da \closedcycle;
    \addplot[forget plot,name path=pr, very thick, sanae5] table [x=t,y=prob,palette-blue] from \da;
    \node[palette-green] (token1) at (3.7, -0.285) {\footnotesize\texttt{\implws of}};
    \node[palette-green] (token2) at (2.5, -0.285) {\footnotesize\texttt{\implws on}};
    \node[palette-green] (token3) at (1.325, -0.32) {\footnotesize\texttt{time}};

    \node[palette-green] (times1) at (3.7, -1.34) {\tiny\color{spearmint} {$\times \p (\mathcal{A} \mid \texttt{pre} \circ \texttt{\implws of}) = \uparrow\uparrow$}};
    \node[palette-green] (times1) at (2.5, -1.302) {\tiny\color{spearmint} {$\times \p (\mathcal{A} \mid \texttt{pre} \circ \texttt{\implws on}) = \uparrow$}} ;
    \node[palette-green] (times1) at (1.325, -1.38) {\tiny\color{spearmint} {$\times \p (\mathcal{A} \mid \texttt{pre} \circ \texttt{time}) = \downarrow\downarrow$}};
    \end{axis}
\end{tikzpicture}
\caption{}
\end{subfigure}%
\begin{subfigure}[t]{0.3\textwidth}
 \centering
     \begin{tikzpicture}
     \hspace{-38pt}
    \begin{axis}[
        const plot,
        width=5cm,
        height=3cm,
        ymax=1,ymin=0,
        xmin=0,xmax=5,
        log origin=infty,
        ticks=none,
        axis lines*=left,
        axis line style={draw=none},
        clip mode=individual,
        rotate around={90:(current axis.origin)},
    ]
    \addplot[forget plot,name path=pr, draw=none, very thick, fill=palette-blue!20!white] table [x=t,y=prob,palette-blue] from \daa \closedcycle;
    \addplot[forget plot,name path=pr, very thick, sanae5] table [x=t,y=prob,palette-blue] from \daa;
    \node[palette-green] (token1) at (3.7, -0.3) {\footnotesize\texttt{\implws of}};
    \node[palette-green] (token2) at (2.5, -0.3) {\footnotesize\texttt{\implws on}};
    \node[palette-green] (token3) at (1.325, -0.34) {\footnotesize\texttt{time}};
    \end{axis}
\end{tikzpicture}
\caption{}
\end{subfigure}%
\end{adjustbox}
\caption{
\textbf{An illustration of our proposed approach.}
(a) Given a prefix, the LM defines a distribution over possible next-tokens.
(b) For each possible next-token, we \emph{efficiently} simulate future generation.
(c) An LM sample induces an approximate LM distribution assigning high probability
to similar samples and low probability to dissimilar samples.
(d) Evaluating a verifier on a single simulated generation, we can use the first-order
information to locally approximate the verifier on \emph{all} possible generations,
factoring in the probability of each generations \wrt the LM.
(e) This yields a probability of the constraint, $\mathcal{A}$, the set of all generations
satisfying a target attributed $\attrib$ being satisfied, used to reweigh the next-token distribution.
(f) This results in a new distribution that discounts fluent but constraint violating generations
in favor of less likely but constraint satisfying generations.
}
\label{fig:overview}
\end{figure*}

In this work, in a departure from the aforementioned approaches, we propose
performing \emph{exact inference in an approximate model}~\citep{koller2009probabilistic}.
We propose {\color{spearmint}S}emantic {\color{palette-blue}Con}trol {\color{violett}E}stimator, or \oursc, which leverages the gradient information of a verifier to tractably perform exact inference over \emph{all} generations satisfying the constraint, allowing precise steering of LM generations by reweighing each probable next
token according to its probability of satisfying the constraint.
More precisely, starting from a \emph{lookahead} sample, we construct a \emph{local, contextualized}
LM distribution that assigns a higher probability to semantically similar sentences and a lower probability to semantically dissimilar ones.
We will show that we can \emph{tractably} and efficiently compute the expected
embedding of \emph{all} sentences \wrt this approximate LM distribution.
Computing the expected embedding allows us to estimate the \emph{expected
probability of the constraint} using a single LM sample and a
single evaluation of the verifier by distributing first-order information
regarding the verifier over the expected embedding.
The next-token distribution is then reweighed by \emph{expected probability
of the constraint} and renormalized to obtain the (approximately) correct
\emph{constrained} next-token distribution. %
Computationally, the expected embedding can be computed in $O(1)$ vectorized time,
whereas the lookahead sample can be drawn efficiently by utilizing an auxiliary model\footnote{We made use of ModernBERT~\citep{modernbert} in our experiments}
to unmask future tokens paired with HogWild! (asynchronous) Gibbs sampling~\citep{
HogwildSGD,HogwildGibbs}, with the synchronization frequency trading off accuracy for
efficiency.
An overview of our approach is in~\cref{fig:overview}.

We evaluated our proposed approach on the tasks of controlling the toxicity and sentiment of LM generations, as well as on controlling the topic of generations.
We observed that our approach was far more likely to satisfy the constraint compared to previous approaches, without compromising the quality of the LM
generations, as measured by perplexity. Our proposed method is an inference-time approach, requiring no data and no fine-tuning, and can be easily integrated
with many previous approaches that enforce syntactic constraints.\footnote{Our code and scripts to reproduce all numbers are publicly available in our GitHub repository.}

\textbf{Contributions\;\;\;}
In summary, we introduce \oursc, an approach that leverages exact probabilistic inference in an approximate model
to exert semantic control over LM generations.
Using a single LM sample coupled with a single verifier evaluation, used to obtain first-order information about
the verifier, we are able to compute an estimate of the probability of the constraint \wrt \emph{all} sentences
in the neighborhood of the LM sample.
\oursc can therefore be seen as a seamless marriage between sampling and exact inference.
Our experiments show that \oursc greatly amplifies an LM's ability to conform to semantic constraints
defined using potential functions while retaining the LM's language modeling capabilities.

\section{Levels of Control: From Syntactic to Semantic Constraints}
\label{sec:definition}

We denote an LM generation of arbitrary length $T$ as $\rvy_{1:T} := \left[ \ry_1 \ry_2 ... \ry_T \right]$,
where $\ry_i$ is the instantiation of random variable $Y_i$ and takes values from a fixed vocabulary 
$\sV = \{1, ..., V\}$.

An LM generation can be subject to one of two types of constraints: syntactic and semantic.
Syntactic (or \textit{lexical}) constraints comprise sets of rules, typically expressed using logical connectives or in some formal language, that restrict the set of permissible values assumed by a random variable $Y_i$ such that there exists some completion $\rvy_{>i}$ of the sentence that satisfies the syntactic constraint $\beta$, given the current prefix $\yprefix$, or to state it more formally
\begin{equation}\label{eq:exists_cont}
    \exists \rvy_{>i}\,\conditional{\beta}{\yprefix}
\end{equation}
An example of such constraint could be a simple logical sentence that disallows an expression deemed
inappropriate to appear as part of an LM's generation, \eg $\lnot(y_i = \text{``full''} \wedge y_{i+1} = \text{``of''} \wedge y_{i+2} = \text{``sh!t''})$~\citep{ahmed2023a}.
Syntactic constraints offer an attractive opportunity for parallelization: we are able to \emph{compile}
syntactic constraints into computational graphs that reuse solutions to subproblems to efficiently
capture the space of all satisfying assignments.
Traversing these computation graphs amounts to efficient parallel evaluation across an exponential
number of possible continuations~\citep{choi2020pc, VergariNeurIPS21}, enabling us to tractably
compute the quantity of interest in~\cref{eq:exists_cont}.

Semantic (or \textit{non-lexical}) constraints, on the other hand, presuppose that LM generations satisfy
certain \textit{attributes} (\eg toxicity, politeness, or positive sentiment).
Such attributes are often hard to ascertain lexically, or in terms of surface-level
features that can be captured using a formal language, \eg ``he's got some attitude!''
invokes a snarky tone that is hard to attribute to any particular token in the generation.
Rather, given a target attribute $a$, we suppose access to a \emph{sequence-level verifier
for $\attrib$}, which we denote by $\phiattr$, that given a sequence $\fully$ assigns a binary value,
either $0$ or $1$, to the sequence $\fully$, \ie $\phiattr(y_{1:T})) \in \{0,1\}$.
We can then define ${\cal{A}}$ as the set of \emph{all} sequences
$\fully$ that satisfy the attribute $\attrib$, \ie ${\cal{A}}\coloneq \{\fully \mid \phiattr(\fully) = 1\}$.
Unlike syntactic constraints, semantic constraints, often implemented as complex
neural networks, are not amenable to the form of compilation that enables us to
efficiently capture the set of all satisfying assignments.
In fact, compiling even a single neuron is known to be NP-hard~\citep{Shi2020BNN}.
Computing~\cref{eq:exists_cont} would thus require that we enumerate every possible
continuation, score it using the verifier, discard continuations for which
the attribute does not hold and renormalize, which is intractable.

\textbf{Prologue. }
In what follows we will relax the verifier $\phiattr$ for an attribute $\attrib$ to be probabilistic.
We will then frame the problem of semantic control as a probabilistic inference problem where we are interested in the posterior LM distribution subject to a semantic constraint.
We will show that the problem can be reduced to that of computing expectations, which we then show how to estimate by performing exact and efficient probabilistic inference in an approximate LM induced by a singular model sample and a single evaluation of the verifier.

\section{Great Expectations}
\label{sec:computing-expectations}
We start by assuming access to the LM distribution, denoted by $\p$,
a sequence-level verifier $\phiattr$ for attribute $\attrib$, and a
prefix $\rvy_{1:i}$ where each token $\ry_j$ assumes values in vocabulary
$\sV$.
Our goal is then to sample from the LM distribution $\p$ a generation
$\rvy_{i+1:T}$ subject to the constraint that the attribute $\attrib$
holds on the entire sequence \ie $\phiattr(\rvy_{1:i} \circ \rvy_{i+1:T})
\in \{0,1\}$. %
That entails sampling a generation that fulfills two distinct desiderata: we expect
the generation to be linguistically sound, or fluent as measured by a model's perplexity, \emph{and} to satisfy attribute $\attrib$.
That is, we are interested in sampling from the LLM distribution conditioned
on the event that the sample belongs to the set of \emph{all} sequences
$\fully$ that satisfy the attribute $\attrib$, which we denote by $\mathcal{A}
\coloneqq \{\fully \mid \phiattr(\fully) = 1\}$.
We can then write the target sampling distribution as
\begin{equation}\label{eqn:problem-statement}
\p(\rvy_{i+1:T} \mid \mathcal{A}, \rvy_{1:i}) 
    \overset{(a)}{=} \frac{\p(\rvy_{i+1:T}, \mathcal{A}\mid\rvy_{1:i})}{\p(\mathcal{A} \mid \rvy_{1:i})} 
    \overset{(b)}{=} \frac{\p(\rvy_{i+1:T}, \mid \rvy_{1:i}) \cdot \phiattr(\yprefix \circ \continuation)}{\sum_{\rvy_{i+1:T}}\p(\rvy_{i+1:T} \mid \rvy_{1:i}) \cdot \phiattr(\yprefix \circ \continuation)},
\end{equation}
where equality $(a)$ follows by the definition of conditional probability,
and equality $(b)$ follows by the definition of marginal probability.
Intuitively,~\cref{eqn:problem-statement} gives us a simple, albeit impractical,
recipe for sampling from the LM distribution conditioned on attribute $\attrib$:
we enumerate all possible generations given the prefix, zeroing out all generations
that violate $\attrib$ according to $\phiattr$, followed by renormalization.
In practice, for a given input $\fully$ and attribute $\attrib$, there is some
\emph{uncertainty} associated with $\phiattr(\fully)$.
That is, we will assume access to a model's estimate $\p( \phiattr(\fully) = 1) \in [0,1]$ of whether $\fully$ satisfies attribute $\attrib$. 
Consequently, in a slight abuse of notation, we will redefine
$\phiattr(\cdot)$ to be $\p( \phiattr(\fully) = 1)$, which should
henceforth be thought of as a \emph{probabilistic} verifier for the attribute $\attrib$.
Under this new definition of $\phiattr(\cdot)$, ~\cref{eqn:problem-statement} can be seen
as reweighing each continuation with the probability of satisfying attribute $\attrib$,
followed by renormalizing the distribution.

State-of-the-art LMs, such as Llama 3~\citep{grattafiori2024llama3herdmodels} and GPT-4~\citep{openai2024gpt4}) are autoregressive, so it is useful to rewrite~\cref{eqn:problem-statement} in terms of the next tokens,
\begin{align}%
 \p(\ry_{i+1} \mid \mathcal{A}, \rvy_{1:i})
    &= \frac{\p(\ry_{i+1} \mid \yprefix) \cdot \p(\mathcal{A} \mid \yprefix \circ \ry_{i+1})}{\p(\mathcal{A} \mid \yprefix)}\label{eqn:psa-1}\\
    &= \frac{\p(\ry_{i+1} \mid \rvy_{1:i}) \mathbb{E}_{\p(.\mid \rvy_{1:i+1})}\left[\phiattr(\yprefix \circ \continuation)\right]}{\mathbb{E}_{\p(. \mid \rvy_{1:i})} \left[\phiattr(\yprefix \circ \continuation)\right]}\label{eqn:psa-2},
\end{align}
where~\cref{eqn:psa-1} follows by the definition of conditional probability and~\cref{eqn:psa-2} follows by the definition of marginal probability and expectations.
It is important to note that, since $\mathcal{A}$ is defined as the set of all sequences
$\fully$ that satisfy $\attrib$, the expectations, both in the numerator and in the denominator
range over sequences of length $T$, requiring that we marginalize over all future continuations
of length $T-i$ and $T-(i+1)$, respectively.
Intuitively, at every generation step we need to ``look ahead'' to determine the probability
that the constraint is violated given the current choice of next token.
If the probability is high, we discount the current choice, and if it is low,
then we reinforce the current choice.
Previous methods have approached this intractable expectation by either learning look-ahead functions parameterized by neural networks, or by sampling.
Next, we will show how to compute the above expectation in closed form by relaxing the target distribution.
\begin{figure*}[!t]
 \centering
\begin{minipage}{0.39\textwidth}

\scalebox{.73}{
\begin{subfigure}[b]{0.4\textwidth}
\centering
\begin{align*}
&\text{sample } \rvs: \texttt{[he's, \implws full, \implws of, \implws shit]} \sim \tilde{\p}(\rvy \mid \texttt{I, \implws think})\\
&\begin{array}{l>{\columncolor{pastel-red!20}}r:>{\columncolor{spearmint!20}}r:>{\columncolor{palette-blue!20!white}}r:>{\columncolor{violett}}r} 
&\texttt{he's: 0.5} & \texttt{full: 0.3} & \texttt{of: 0.8} & \texttt{shit: 0.4}\\ %
              &\texttt{it's: 0.25}& \texttt{made: 0.5} & \texttt{on: 0.1} & \texttt{crap: 0.4}\\
              &\texttt{she's: 0.25}& \texttt{smell: 0.2} & \texttt{from: 0.1} & \texttt{hate: 0.2}
\end{array}\\
\end{align*}
\end{subfigure}
}
\begin{subfigure}[t]{0.5\textwidth}
 \centering
 \includegraphics[page=1, scale=0.8, angle=90,origin=c, trim=0 -30pt 0 0]{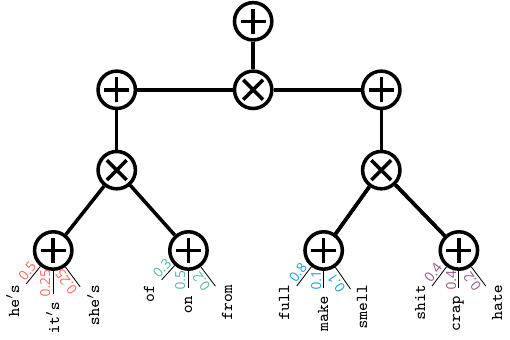}
\end{subfigure}%
\end{minipage}
\begin{minipage}{0.60\textwidth}
\hspace{38pt}
\begin{subfigure}[t]{0.5\textwidth}
 \centering
 \vspace{-28pt}
 \includegraphics[page=2, scale=0.8,angle=90,origin=c]{figures/semprola.pdf}
\end{subfigure}%
\end{minipage}
\vspace{-30pt}
\caption{
\textbf{A technical overview of our approach.}
(top left) We start by sampling an approximate generation $\rvs$ using Gibbs
sampling $\tilde{p}$ conditioned on the prefix from the model's marginal
conditionals, $\p(\rvy_{i} \mid \rvy_{-i}) \forall_i$.
Conditioned on $\rvs$, the models marginal conditionals induce a distribution
on all generations, assigning higher probabilities to similar sentences and
lower probabilities for dissimilar sentences, which we visualize for the top-3
tokens for clarity of exposition.
(bottom left) We can parameterize a \emph{circuit} using the above distribution,
yielding a closed-form, tractable representation of probability distribution
defined in~\cref{eq:local_psl}, where read left to right, every leaf node corresponds
to a categorical distribution on $\rvy_i$
(right) Such a representation enables us to compute the expected embeddings \wrt the
distribution in the neighborhood of the sample $\rvs$ by substituting token embedding
for corresponding embeddings at leaf nodes, computing weighted sums of embeddings at 
sum nodes, and taking sums at product nodes.
This allows us to plug the expected embedding into \cref{eqn:tractable-taylor-approximation}
to yield the constraint probability.%
}
\label{fig:circuit}
\end{figure*}

\section{Semantic Probabilistic Control}%
\label{sec:method}

The computational hardness of the expectations that we introduced in~\cref{eqn:psa-2} can intuitively be attributed to the \emph{lack of structure} along two distinct dimensions.

First, is the \emph{lack of structure to the distribution}.
Consider computing the probability that a sequence of length $T$ ends in the word ``love''.
Computing such a probability under the autoregressive distribution requires that we marginalize over all possible sequences ending in ``love'', roughly $O(|\sV|^T)$.
In fact, computing such probability is known to be computationally intractable~\citep{Roth93}.
Contrast that with a fully-independent\footnote{where $\p(\fully) = \prod_{i=1}^{T} \p(\ry_{i})$, \ie the probability of a token is independent from all other tokens.} distribution, where we can simply query the network for the probability of a given
token in constant time.
Clearly there is a tension here: fully-independent distributions, while easier to
reason about, are not expressive and therefore do not make for good LMs, whereas autoregressive distributions are harder to reason about, but a lot more expressive, and achieve SoTA language modeling.

The second dimension is the \emph{lack of structure to the constraint}.
Recall that we have assumed $\phiattr$ to be a neural network, which prior work has shown to be computationally intractable to decompose over sequences~\citep{Shi2020BNN}~\footnote{in fact, the problem remains intractable even assuming $\phiattr$ is a single neuron~\citep{Khosravi2019expect}.}.
That is, given $\phiattr(\yprefix)$ for a prefix $\yprefix$, we know of no way of efficiently
extending $\phiattr(\yprefix)$ to $\phiattr(\yprefix \circ \ry_{i+1})$ by only processing
the new element $\ry_{i+1}$ and reusing the result of the previous evaluation $\phiattr(\yprefix)$.

\subsection{Locally Contextualized Distribution}
To sidestep the hardness of the autoregressive distribution, we move towards the tractability of fully-independent distributions, while retaining as much of the contextual information.
Therefore, we consider the \emph{pseudolikelihood} of a sentence~\citep{Besag1975, ahmed2023a},
\begin{equation}\label{eq:pseudo_likelihood}
    \p(\fully) \approx \Tilde{\p}(\fully) \coloneqq \prod_{i}\p(\rvy_i \mid \rvy_{-i}),
\end{equation}
where $\rvy_{-i}$ denotes $\rvy_1, \ldots, \rvy_{i-1}, \rvy_{i+1}, \ldots, \rvy_n$.
Unfortunately, \cref{eq:pseudo_likelihood} does not ensure tractability, seeing that different sentences would depend on different sets of conditionals.
We define the pseudolikelihood of a sentence $\rvy$ \emph{in the semantic neighborhood of a sentence}~$\ytilde$
\begin{equation}
    \Tilde{\p}_{\ytilde}(\rvy) \coloneqq \prod_{i}\p(\rvy_i \mid \ytilde_{-i}) \label{eq:local_psl}
\end{equation}
which can be thought of as the \emph{contextualized probability} of a sentence $\rvy$ given the context $\ytilde$.
That is, \cref{eq:local_psl} calculates the probability of sequence $\rvy$ by taking the product of probabilities of each token $\rvy_i$, crucially conditioning each token $\rvy_i$ not on the preceding tokens of $\rvy$, but on the context surrounding position $i$ within $\ytilde$ (specifically, $\ytilde$ excluding its i-th token, denoted $\ytilde_{-i}$).
Therefore, $\ytilde$ acts as a contextual anchor for evaluating $\rvy$ under this measure. Intuitively, sentences y that semantically or structurally align well with the specific token-level contexts provided by $\ytilde$ are expected to yield a higher pseudolikelihood~$\Tilde{\p}_{\ytilde}(\rvy)$.

\begin{figure}[tb]
\begin{adjustbox}{minipage=\linewidth,scale=0.96}
 \begin{minipage}[t]{\dimexpr.53\textwidth-.5\columnsep}
    \begin{algorithm}[H]
    \begin{algorithmic}[1]
    \STATE\textbf{Input}: Verifier $\phiattr$, LM distribution\\ $\p(\ry_i \mid \yprefix)$, prefix $\rvy_{1:i}$, max length $T$\\\vspace{2pt}
    \STATE\textbf{Output}: $\p(\ry_{i+1} \mid \rvy_{1:i}, \mathbb{A})$\\\vspace{2pt}
    \COMMENT\hspace{-11pt}\textcolor{spearmint}{$\triangleright$ Expand the batch to include top-k tokens}\\\vspace{2pt}
    \STATE $\mathtt{top_k} = \argmax_k \p(\ry_i \mid \yprefix)$
    \STATE $\rvy_{1:i+1} = \rvy_{1:i}\mathtt{.expand(n, top_k)}$\\\vspace{2pt}
    \COMMENT\hspace{-11pt}\textcolor{spearmint}{$\triangleright$ Get $N$ samples $\stilde$ from $\p(\rvy_{i+2:T} \mid \rvy_{1:i+1})$}\\\vspace{2pt}
    \STATE $\stilde^1$, \ldots, $\stilde^N \sim \mathtt{GibbsSampler}(\rvy_{1:i+1}, \p)$ \\\vspace{2pt}
    \COMMENT \hspace{-11pt}\textcolor{spearmint}{$\triangleright$ Estimate prob $q$ of satisfying constraint}%
    \STATE $q = \mathtt{zeros}(\mathtt{top}_k) $\\\vspace{2pt}
    \FOR{each $\tilde{\rvs}$ in $\stilde^1$, \ldots, $\stilde^N$}
        \STATE $\condmarg = \mathtt{CondMarginals}(\p, \stilde_{i+2:T})$\vspace{2pt}%
        \STATE $\gradphiattr = \mathtt{LinearizeVerifier}(\phiattr, \stilde)$\vspace{2pt}%
        \STATE $q[\tilde{\rs}_{i+1}]$ += $\mathtt{EstimateProb}(\condmarg, \phiattr, \gradphiattr)$\vspace{2pt}%
    \ENDFOR\\
    \COMMENT \hspace{-11pt}\textcolor{spearmint}{$\triangleright$ Renormalize $\q$}\\\vspace{2pt}
    \STATE $log \q = \q\mathtt{.log\_sofmax}()$ \\\vspace{2pt}
    \COMMENT \hspace{-11pt}\textcolor{spearmint}{$\triangleright$ Reweight the LM distribution}\\\vspace{2pt}
    \STATE $\mathtt{w} = \log \p(\rvy_{i+1}|\rvy_{1:i}) + \log \q$\\
    \STATE $\p^{*} = \mathtt{Categorical(weights=w)}$\\\vspace{2pt} 
    \STATE $\mathtt{return}$ $\p^{*}$\vspace{0.1em}
\end{algorithmic}
\caption{\oursc\vspace{0.13em}}
\label{algo:proposed-algo}
\end{algorithm}
\end{minipage}
 \begin{minipage}[t]{\dimexpr.53\textwidth-.5\columnsep}
    \begin{algorithm}[H]
    \begin{algorithmic}[1]
    \STATE \textbf{Input}: Verifier $\phiattr$,  Sample $\rvs$ \\\vspace{2pt}
    \STATE \textbf{Output}: Gradient of $\phiattr$ \wrt $\rvs$ embedding \\\vspace{2pt}
    \COMMENT\hspace{-11pt}\textcolor{spearmint}{$\triangleright$ Obtain embeddings for $\rvs$}\\
    \STATE $\mathtt{emb\_layer} = \phiattr.\mathtt{get\_input\_embeddings}()$ \\
    \STATE $\mathtt{emb} = \mathtt{emb\_layer}(\rvs)$\\
    \COMMENT\hspace{-11pt}\textcolor{spearmint}{$\triangleright$ Collect gradient of $\phiattr$ \wrt to $\mathtt{emb}$}\\
    \STATE $\mathtt{score} = \phiattr(\mathtt{emb})\mathtt{.sum()}$ \\
    \STATE $\mathtt{grad} = \mathtt{autograd.grad}(\mathtt{score}, \mathtt{emb})$\\
    \STATE $\mathtt{return}$ $\mathtt{grad}$\vspace{0.1em}
\end{algorithmic}%
\caption{$\mathtt{LinearizeVerifier}$}\vspace{0.13em}
\label{algo:gradient-approximation}
\end{algorithm}

 \vspace{-2.30em}
\begin{algorithm}[H]
\begin{algorithmic}[1]
    \STATE \textbf{Input}: Conditional marginals $\condmarg$, Verifier $\phiattr$,  Gradient $\nabla_{\mathtt{emb}(\rvs)} \phiattr$, $\mathtt{embs} \coloneq [\overline{\mathtt{emb}}(\rvy_{i,1}), \ldots,  \overline{\mathtt{emb}}(\rvy_{i,|\sV|})]$ , score $\phiattr(\rs)$, $T$ \\\vspace{2pt} 
    \STATE \textbf{Output}: $\p(\mathcal{A} \mid \yprefix)$ \\\vspace{2pt}
    \COMMENT\hspace{-11pt}\textcolor{spearmint}{$\triangleright$ Compute expected embedding}\\
    \STATE $\mathtt{exe} = 0$\\ 
    \FOR{i in 1, \ldots, T}
    \STATE \hspace{-0.5em}$\mathtt{exe}$ += $\mathtt{embs}[\ldots, \mathtt{None}] \cdot \condmarg[:, \mathtt{i:i+1},:]$\\
    \ENDFOR
    \STATE $\mathtt{exe = exe.mean(0)}$\\
    \COMMENT\hspace{-11pt}\textcolor{spearmint}{$\triangleright$ First-order Taylor expansion about $\rvs$}\\
    \STATE $\mathtt{return}$ $\phiattr(\rvs) + \nabla_{\mathtt{emb}(\rvs)}\phiattr \cdot (\mathtt{exe} - \overline{\mathtt{emb}}(\rvs))$\vspace{-0.32em}
    
\end{algorithmic}
\caption{$\mathtt{EstimateProb}$}
\label{algo:estimate-prob}
\end{algorithm}
 \end{minipage}
 \end{adjustbox}
\end{figure}
\subsection{Bridging Samples and Expectations: A Tangential View}
\label{ssec:first-order-taylor-approx}
Next, we turn our attention to address the \emph{hardness of the verifier $\phiattr$}.
In particular, given an LM sample $\rvs \sim \p(\rvy_{i+1:T}|\rvy_{1:i})$
and access to a verifier $\phiattr$, we leverage gradient information obtained during the evaluation of $\phiattr(\rvs)$, coupled with
the contextualized probability distribution in \cref{eq:local_psl}, to approximate $\mathbb{E}_{\p(. \mid \rvy_{1:i})} \left[\phiattr(\fully)\right]$, the constraint probability.

We denote by $\mathtt{emb}: \sV \mapsto \mathbb{R}^d$ an embedding function that maps each token onto a $d$-dimension vector and let  $\overline{\mathtt{emb}}(\rvy)$ denote the average token-wise embedding.\footnote{\wolog, we assume this embedding can be extracted directly from the embedding layer of the verifier, \ie $\phiattr(\rvs) \coloneqq \phiattr(\mathtt{emb}(\rs_1), \cdots, \mathtt{emb}(\rs_T))$.} 
Then, we can approximate \cref{eqn:psa-2} using a first-order Taylor expansion of $\phiattr$ about the LM sample $\rvs$
\begin{equation}\label{eqn:taylor-approximation}
\mathbb{E}_{\Tilde{\p}(\cdot \mid \rvy_{1:i})} \left[ \phiattr(\rvy_{1:T}) \right] \approx \mathbb{E}_{\Tilde{\p}(\cdot \mid \rvy_{1:i})} \left[  \phiattr(\rvs) + \nabla \phiattr(\rvs) \cdot (\overline{\mathtt{emb}}(\rvy_{1:T})-\overline{\mathtt{emb}}(\rvs)) \right].
\end{equation}
Using the linearity of expectation, we can further simplify expression, obtaining
\begin{equation}\label{eqn:tractable-taylor-approximation}
\mathbb{E}_{\Tilde{\p}(\cdot \mid \rvy_{1:i})} \left[ \phiattr(\rvy_{1:T}) \right] \approx  \phiattr(\rvs) + \nabla \phiattr(\rvs) \cdot (\mathbb{E}_{\Tilde{\p}(\cdot \mid \rvy_{1:i})} \left[ \overline{\mathtt{emb}}(\rvy_{1:T}) \right] -\overline{\mathtt{emb}}(\rvs)). 
\end{equation}
We have now managed to \textbf{\emph{reduce the problem of estimating the constraint probability}}, given by the expectations in~\cref{eqn:psa-2} \textbf{\emph{to the problem
of computing an average sentence embedding}} \wrt an approximate LM distribution
$\tilde{\p}$, followed by simple arithmetic operations.
We will next show how we can efficiently compute the expected sentence embedding.

\subsection{From Sequence Probabilities to Average Embeddings}
\label{ssec:tractable_expectation}
We appeal to knowledge compilation, a class of methods that transform, or \emph{compile},
a function into a tractable target form which represents functions as parameterized
computational graphs, or \emph{circuits}.
By enforcing certain structural properties on the compiled circuits, we can enable the
tractable computation of corresponding classes of probabilistic queries.
Thus, circuits provide a language for constructing and reasoning about tractable
representations.

Formally, a \emph{circuit} $\p$ over variables $\rvY$ is a parameterized computational
graph encoding a function $\p(\rvY)$.
Each node $n$ in the graph encodes a parameterized function $\p_n(\vars(n))$ over
variables $\vars(n) \subseteq \rvY$, also known as its \emph{scope}.
Each inner node in the graph is a sum or a product node, and each leaf node encodes
a tractable input distribution over its scope. 
Each inner unit $n$ (\ie product or sum node) receives inputs from other units, denoted $\ch(n)$.

A circuit is \emph{decomposable} if the inputs of every product node depends on
disjoint sets of variables, \ie for $n = c_1 \otimes c_2$, $\vars
(c_1) \cap \vars(c_2) = \varnothing$.
Intuitively, decomposable product nodes encode local factorizations over
variables of the function.
We assume that decomposable product nodes always have two inputs, a condition
that is enforceable on any circuit in exchange for a polynomial increase in its size~\citep{vergari2015simplifying,peharz2020einsum}.

A second property is \emph{smoothness}.
A circuit is \emph{smooth} if the inputs
of every sum node depend on the same set of variables, \ie for
$n = \bigoplus_i \theta_i \cdot c_i$, $\vars(c_i) = 
\vars(c_j)\ \forall i,j$. Decomposability and smoothness
are sufficient and necessary for tractable integration
over arbitrary sets of variables in a single pass, as they allow
larger integrals to decompose into smaller~ones.
Given a circuit for a distribution $\tilde{\p}$, the expected embedding can then be computed by traversing the circuit bottom-up,
substituting token embedding for corresponding embeddings at leaf nodes, computing weighted sums of embeddings at sum nodes, and
taking sums (in essence, concatenating embeddings) at product nodes, as can be seen in~\cref{fig:circuit}.

\subsection{Closing the Loop}
\label{ssec:probability-weighting}
Our full algorithm is given in \cref{algo:proposed-algo}.
We start by truncating the next-token distribution using top-k or top-p,
as is common place in modern autoregressive LMs, where we use top-k for clarity of exposition.
We then proceed by simulating a continuation for each of the possible top-k tokens, each produced using a
masked LM and Hogwild! Gibbs sampling\footnote{We refer the reader to~\cref{sec:Gibbs} for more details.}, to avoid
expensive autoregressive sampling from the LM. 
We then proceed by computing the contextualized probability of each sample $\sV_i$ and the gradient of the
verifier \wrt the sample embedding $\nabla_{\mathtt{emb}(\rvs)} \phiattr$, used to estimate the constraint
probability.
Having computed the constraint probability, we reweigh the next-token distribution to account
for the constraint being satisfied, and renormalize to obtain the conditional next-token distribution.

\section{Related Work}
\label{sec:related_work}

Recent advances in controllable generation with LMs have spurred a wide range of approaches, which we summarize below. %
These approaches can be roughly classified into three different categories: \textit{training-time}, \textit{prompting}, and \textit{decoding-time} approaches.

\textbf{Training-time approaches. } A subset of the approaches seeks to exert control by fine-tuning or reinforcement learning via some set of data that more closely mirrors the target task, such as via reinforcement learning from human feedback (RLHF) \citep{ziegler2019fine,stiennon2020learning,bai2022constitutional,ouyang2022training} or from symbolic knowledge~\citep{ahmed2023a}, but these approaches come with challenges such as hyperparameter sensitivity and distributional collapse \citep{zheng2023secrets,zhu2023principled,xiong2024iterative}.
Some of these drawbacks can be mitigated by utilizing on-policy data \citep{tajwar2024preference} and imposing a KL penalty that penalizes shifting an LM too far from its prior distribution, casting optimization as variational inference \citep{korbak2022rl,amini2025variational}.

\textbf{Prompting approaches. }
Another class of approaches focuses on guiding the distribution implicitly via modifications in the prompt~\citep{ashok2024controllabletextgenerationinstructiontuning}. 
To this end, control can be exerted by either verbally expressing the constraints in the prompt~\citep{chen2022controllabletextgenerationlanguage,pmlr-v202-zhou23g,ashok2024controllabletextgenerationinstructiontuning}, or through the use of examples~\citep{poesia2022synchromesh,pmlr-v202-zhou23g}. 
In addition to introducing minimal computation overhead and producing good quality text~\citep{pmlr-v202-zhou23g,ashok2024controllabletextgenerationinstructiontuning}, prompting approaches are also more flexible, since complex constraints can be easily integrated in the prompt without further training or expensive data curation. 
Nonetheless, constraint satisfiability using prompting-based methods is not guaranteed~\citep{pmlr-v202-zhou23g} and depends heavily on the instruction following capabilities of the LM~\citep{jiang-etal-2024-followbench,he-etal-2024-complex}.

\textbf{Decoding-time approaches. }
A popular decoding-time approach is to perform token-level modifications at each step and, for that reason, frequently referred to as \textit{locally constrained decoding}~\citep{loula2025syntactic-2025-ICLR}.
Methods to locally constrained decoding either mask out specific tokens or heuristically reweigh tokens such that the constraints are more likely to be satisfied.
Examples include 
banning specific words~\citep{gehman-etal-2020-realtoxicityprompts}, 
using context-free grammars~\citep{poesia2022synchromesh,geng-etal-2023-grammar,outlines,Beurer-Kellner2023,guidance,Beurer-Kellner-ICML2024-DOMINO}, or through the combination of boolean algebra with search algorithms~\citep{hokamp-liu-2017-lexically,anderson-etal-2017-guided,post-vilar-2018-fast,hu-etal-2019-improved,lu-etal-2021-neurologic,lu-etal-2022-neurologic,qin2022cold}. 
Note, however, that while setting token-level restrictions can be effective at exerting syntactic control over LMs, these are insufficient to capture the richer and subtler nuances of semantic constraints.

In fact, semantic control approaches resort to attribute ``scorers'' to estimate how likely the constraint is under a given input, and then use those estimates to reweigh the per-token distribution of the base LM. 
Previously proposed methods include
combining the conditional distributions of different LMs with opposing behaviors, such as a toxic expert and a non-toxic expert~\citep{schick-etal-2021-self,liu-etal-2021-dexperts,li-etal-2023-contrastive,dekoninck2024controlled}, and 
using an attribute discriminator (\ie constraint verifier) to reweigh the base LM conditional distribution~\citep{holtzman-etal-2018-learning,krause-etal-2021-gedi-generative}.
The gradients of attribute discriminators have also been to induce changes the base LM through changes to the LM weights~\citep{Dathathri2020Plug,liu-etal-2020-data,wallace-etal-2019-universal,zhang-etal-2024-p4}.
Although effective, locally constrained decoding approaches often introduce greedy (potentially sub-optimal) approximations that distort the distribution~\citep{loula2025syntactic-2025-ICLR,ma2025nonmyopic}.  
Conversely, sample-reweigh approaches consist of first sampling complete sequences and then reweigh them using a constraint verifier~\citep{Stiennon2020-BestOfN-2020-Neurips,krishna-etal-2022-rankgen,sun2024fastbestofndecodingspeculative,ichihara2025evaluation-2025-TMLR,amini2025variational}. 
While constraints are imposed globally in sample reweighing approaches, they do not benefit from finer-grained constraint information during generation and, hence, require a larger number of samples to find high-quality generations that comply with the constraints~\citep{loula2025syntactic-2025-ICLR}. %

Another line of work performs approximate inference in exact models via sampling~\citep{Miao2019,zhang-etal-2020-language-generation,kumar-etal-2022-gradient,poesia2022synchromesh,qin2022cold,pmlr-v235-du24a}, and, more recently, via more effective Sequential Monte Carlo (SMC) methods, which maintain a set of samples that evolve through time. 
The evolution of the samples accounts not only for the sample likelihood under the base LM, but also for constraint information that can be provided either by learnable twist functions~\citep{Zhao-et-al-2024-twist-functions-2024-ICML} or by evaluating the constraint verifier on partial sequences ~\citep{lew2023sequentialmontecarlosteering,loula2025syntactic-2025-ICLR}.

\section{Experiments}
\label{sec:experiments}

We empirically evaluate the effectiveness of the proposed method across numerous open-ended generation tasks, including text detoxification, controlled sentiment generation, and topic steering. 
Section \ref{ssec:experiment-setup} introduces specific details of our methods, baselines, and metrics.
Task-specific details, such as dataset and constraint verifiers, and results for the toxicity, sentiment, and topic experiments are described in Sections \ref{ssec:toxicity-control-experiments}, \ref{ssec:sentiment-control-experiments}, \ref{app:additional-results-topic-control}, respectively.
\begin{table}[tb]
\centering
\caption{\textbf{Evaluation of the quality and toxicity of \llama generations when steered to be {\color{SeaGreen}non-toxic} and \color{WildStrawberry}toxic}, respectively. 
Toxicity is evaluated on 400 prompts \realtoxicityprompts using the toxicity verifier $\phi_\mathtt{toxicity}$~\citep{logacheva-etal-2022-paradetox}. 
 \textbf{PPL} refers to the perplexity of \perplexitymodel on the model generations.
We report \textbf{Expected Maximum Toxicity}: the maximum toxicity across generations, and \textbf{Toxicity Probability}: the probability of a toxic generation, both computed across 10 generations per prompt. 
We expect both metrics to be lower ({\color{SeaGreen}$\downarrow$}) when steering the base LM towards non-toxic generations ({\color{SeaGreen}\textbf{detoxify}}) and higher ({\color{WildStrawberry}$\uparrow$}) when steering the base LM towards non-toxic generations ({\color{WildStrawberry}\textbf{toxify}}).
}
\label{tab:experiments-detoxification-results}
\resizebox{\columnwidth}{!}{%
\begin{tabular}{ll ccc ccc c}
\toprule
\multicolumn{1}{l}{\multirow{2}{*}{\textbf{Objective}}} &\multicolumn{1}{l}{\multirow{2}{*}{\textbf{Method}}} & \multicolumn{3}{c}{\textbf{Toxic Prob. $({\color{SeaGreen}\downarrow}, {\color{WildStrawberry}\uparrow})$}}  & \multicolumn{3}{c}{\textbf{Exp. Max. Toxicity $({\color{SeaGreen}\downarrow}, {\color{WildStrawberry}\uparrow})$}} & \multicolumn{1}{c}{\multirow{2}{*}{\textbf{PPL $(\downarrow)$}}}\\
\cmidrule(lr){3-5}
\cmidrule(lr){6-8}
 && \textbf{Full} & \textbf{Non-toxic} & \textbf{Toxic} & \textbf{Full} & \textbf{Non-toxic} & \textbf{Toxic} & \\
\midrule
&\random & $37.25$ & $10.00$ & $64.50$ & $37.11$ & $13.17$ & $61.05$ & $12.18$ \\
&\beamsearch & $17.25$    &   $3.00$ & $31.50$ & $18.22$ & $4.34$ & $32.09$ &  $\mathbf{8.00}$  \\ 
\midrule\midrule
\multicolumn{1}{l}{\multirow{2}{*}{\color{SeaGreen}\textbf{detoxify}}} &\bestofn   & $2.75$	& $1.00$	 & $4.50$	& $4.90$	& $1.91$	& $7.89$ & $15.46$ \\
&\ours (ours) & $\mathbf{00.25}$ & $\mathbf{00.50}$ & $\mathbf{00.00}$ &  $\mathbf{01.85}$ & $\mathbf{1.30}$ &  $\mathbf{2.40}$ & $14.88$ \\
\midrule
\multicolumn{1}{l}{\multirow{2}{*}{\color{WildStrawberry}\textbf{toxify}}} &\bestofn    & $62.50$	& $37.00$	& $88.00$	& $61.36$	& $39.62$	& $83.11$ & $13.97$ \\ %
&\ours (ours)& $\mathbf{93.75}$ & $\mathbf{88.00}$	& $\mathbf{99.50}$	& $\mathbf{91.15}$	& $\mathbf{85.75}$ & $\mathbf{96.55}$ & $23.87$ \\ 
\bottomrule
\end{tabular}
}
\end{table}
\subsection*{Experimental Setup}
\label{ssec:experiment-setup}

\textbf{Baselines. }
To validate our method, we compare it against two sampling-based baselines:
\random, which consists of sampling outputs autoregressively from a base LM, 
and \beamsearch, which leverages information about the top $K$ most likely continuations under a base LM to greedily select the next token. 
Additionally, we evaluate 
\textbf{Best-of-N rejection sampling (\bestofn)} ~\citep{Stiennon2020-BestOfN-2020-Neurips}, a popular training-free method for language model control which has been shown to be competitive to RLHF-based methods~\citep{amini2025variational}.
Like our proposed method, \bestofn exploits non-lexical constraint verifiers to exert semantic control on the base LM. 
However, it does so by first sampling $N$ continuations from the base LM and selecting one that maximizes the verifier.\footnote{For a fair comparison, we use the same decoding settings as in our method's initialization.} 
We refer to Appendix \ref{appdx:additional-details:methods} for more details.

\textbf{Metrics. }
In line with prior work~\citep{gehman-etal-2020-realtoxicityprompts,Ahmed2025controllable}, we report \textbf{Perplexity} (\textbf{PPL}) as a measure of sample quality, specifically, we use \perplexitymodel. 
Intuitively, effective control methods should yield generations that satisfy the contraint but that are also high quality, \ie low perplexity.

The primary constraint satisfaction metric that we report is the \textbf{Average $\phiattr$ score}. This metric can be defined as the average verifier score across all model generations. 
Intuitively, because this verifier is being used to steer control during generation, it can be interpreted as the \textit{ground truth} measure of the desired semantic attribute $\bm{a}$ (\eg toxicity, positive sentiment, topic). 
As such, we expect effective control methods to achieve high Average $\phiattr$ scores, especially when compared to uncontrolled baselines like \random. 

As additional measures of constraint satisfaction, we report metrics that capture the expected worst-case and the empirical probability of constraint satisfaction~\citep{gehman-etal-2020-realtoxicityprompts}. 
Assuming that each prompt $x$ is associated with multiple generations, the \textbf{expected worst score metric} is calculated by computing the worst constraint score $\phiattr$ across all generations for $x$, and, then taking the average over all evaluation prompts.
Similarly, the \textbf{constraint probability} metric represents the fraction of evaluated prompts for which at least one of its generations satisfies the constraint above a user-defined threshold (\ie $\mathbb{1}[\phiattr(\rvy) \geq \tau_{\bm{a}}]$).

\begin{table}[t!]
\small
\centering
\caption{\textbf{Evaluation of quality and sentiment of \gptimdb generations when steered using a positive sentiment constraint \phisentiment}. 
Sentiment is evaluated on 600 prompts from the \imdb test set using a sentiment verifier~\citep{maas-etal-2011-imdb-dataset}, spanning equal number of positive and negative reviews.
Results are discriminated by the \textbf{Full} set of prompts, the \textbf{Neg}ative subset, and the \textbf{Pos}itive subset.
All metrics are calculated using 10 different generations per prompt.
\textbf{PPL} refers to the perplexity of \perplexitymodel on the model generations using 10 different seeds;
In line with \citet{rafailov2023direct,amini2025variational}, we report the average sentiment score, the sentiment score is greater than 0.8 in 9 out 10 generations (\textbf{Sentiment Prob.}), and the expected minimum sentiment score (\textbf{Exp. Min. Sentiment}). 
}
\resizebox{\columnwidth}{!}{%
\begin{tabular}{l ccc ccc ccc c}
\toprule
\multicolumn{1}{l}{} 
    & \multicolumn{3}{c}{\textbf{Avg \phisentiment} ($\uparrow$)}
    & \multicolumn{3}{c}{\textbf{Sentiment Prob.} ($\uparrow$)}  
    & \multicolumn{3}{c}{\textbf{Exp. Min. Sentiment} $(\uparrow)$} 
    & \multicolumn{1}{c}{\textbf{PPL $(\downarrow)$}}\\
\cmidrule(lr){2-4}
\cmidrule(lr){5-7}
\cmidrule(lr){8-10}
\textbf{Method} 
    & \textbf{Full} & \textbf{Neg} & \textbf{Pos} 
    & \textbf{Full} & \textbf{Neg} & \textbf{Pos} 
    & \textbf{Full} & \textbf{Neg} & \textbf{Pos} & \textbf{Full} \\
\midrule
\random     & $57.10$	& $53.16$ & $61.04$ & $95.50$ & $ 	95.33$ & $ 	 95.67$ & $12.83$ &   $10.78$ & $ 	14.87$ & $ 	21.18$ \\
\beamsearch & $58.83$	& $50.83$ &	$66.82$ & $58.83$ & $ 	48.33$ & $ 	69.33$ & $ 	44.46$ & $ 	37.21$ & $ 	51.71$ & $ 	\mathbf{3.96}$ \\ 
\bestofn  & 60.66	& 55.17 &	66.14  &  $95.83$ & $ 	93.33$ & $ 	98.33$ & $ 	15.24$ & $ 	11.70$ & $ 	18.77$ & $ 	10.84$ \\ 
\ours (ours)    & $\mathbf{93.06}$  & $\mathbf{92.73}$ &	$\mathbf{93.37}$   & $\mathbf{100.00}$ & $\mathbf{ 	100.00}$ & $ 	\mathbf{100.00}$ & $ 	\mathbf{84.50}$ & $ 	\mathbf{83.18}$ & $ 	\mathbf{85.82}$ & $ 	20.96$ \\ 
\bottomrule
\label{tab:sentiment-exps-break-down-by-prompt-sentiment}
\end{tabular}
}
\end{table}
\subsection{Controlled Toxicity Generation}
\label{ssec:toxicity-control-experiments}

In this section, we compare the performance of different methods in steering the toxicity of a small \llama~\citep{grattafiori2024llama3herdmodels}. 
We do so by prompting the LM with 400 natural occurring prompts from \realtoxicityprompts~\citep{gehman-etal-2020-realtoxicityprompts}.
We randomly select 200 \emph{Toxic} and 200 \emph{Non-toxic} prompts from \realtoxicityprompts and use them in both toxification and detoxification settings, sampling 10 generations of up to 25 tokens per prompt.
Evaluation and toxicity steerability are both conducted using a RoBERTa-based binary classifier \phitoxicity, finetuned for toxicity detection~\citep{logacheva-etal-2022-paradetox}.
To steer models to generate non-toxic outputs, we set them to maximize $1 -\phi_{\mathtt{toxicity}}$.

\textbf{Detoxification Task. } 
Table \ref{tab:experiments-detoxification-results} summarizes the results for the detoxification task, discriminated by prompt type. 
Intuitively, effective semantic control methods should be able to generate non-toxic outputs, \ie minimize the toxicity metrics, irrespective of the toxicity of the prompt type. 
Overall, we observe that the uncontrolled baselines \random and \beamsearch, still lead to toxic continuations even when prompted with non-toxic inputs. 
While \beamsearch seems to lower both toxicity and perplexity, we find that this is explained by degenerate outputs characterized by repetition~\citep{Holtzman2020The}.
Contrastingly, we find that \bestofn is very effective at detoxifying LM generations: reducing the average worst-case toxicity down 4.90 with minimal penalty in perplexity (3.28 points). 
While this represents a big improvement over the uncontrolled baselines, we find that our method is able to further achieve a 3-fold reduction in terms of the average worst case toxicity in toxic prompt and reduce the probability of a toxic generation to a negligible amount (up to 0.50).

\textbf{Toxification Task.} 
We now move to the opposite task: given a naturally occurring prompt, are methods able to steer the base LM towards more toxic inputs?
Table~\ref{ssec:toxicity-control-experiments} shows the toxicity results for the semantic control methods. 
While both methods are able to substantially increase both the worst-case toxicity and the likelihood of sampling toxic outputs from \llama, %
we find that \ours systematically is far more effective than \bestofn with +30\% gap toxicity increase across both toxicity metrics.
Much of this performance gap appears to stem from the non-toxic subset, for which the base LM is less predisposed to generate toxic outputs. 
As such, methods like rejection sampling that use the constraint verifier \phitoxicity to rerank the base LM generations are less likely to succeed for low probability semantic constraints.
This also provides an explanation for the increase in perplexity for \ours.

\subsection{Controlled Sentiment Generation}
\label{ssec:sentiment-control-experiments}

Next, we compare the steerability of the different methods when generating reviews with positive sentiment~\citep{rafailov2023direct,Zhao-et-al-2024-twist-functions-2024-ICML,amini2025variational}.
Focusing on movie reviews, we prompt \gptimdb with 600 arbitrarily chosen prompts from the \imdb  test set~\citep{maas-etal-2011-imdb-dataset}. 
Building on previous work~\citep{rafailov2023direct,amini2025variational}, we use the original reviews in the \imdb dataset to create the prompts by randomly splitting them into prefixes of 2 to 8 words. 
We also adopt the same BERT-based classifier as our sentiment verifier \phisentiment.\footnote{\url{https://huggingface.co/lvwerra/distilbert-imdb}} Given that this model was fine-tuned on the \imdb training data, we expect it to be a strong and reliable sentiment predictor for this task.

\textbf{Positive Movie Review Generation Task. }
In the context of positive movie review generation, we would like to ensure that most of \gptimdb's generations are positive.\footnote{In line with ~\citet{maas-etal-2011-imdb-dataset}, we consider a review to be positive iff \phisentiment$(\rvy)\geq 0.8$.}
Once more, as observed in Table \ref{tab:sentiment-exps-break-down-by-prompt-sentiment}, the uncontrolled baselines\textemdash\random and \texttt{beamsearch}\textemdash struggle to generate positive reviews. 
Specifically, as emphasized by the worst case metric, Expected Minimum Sentiment, \gptimdb-generated reviews with no control can be fairly negative ($<52$ across all prompts), especially in the negative subset ($<38\%$).
\bestofn drastically improves upon the uncontrolled baselines, increasing the Sentiment Probability to about 70.83\% and improving the average lowest sentiment score to 70.79\%. 
Still, we find that \ours further improves (about 14\% points average improvement in both metrics) the overall worst-case sentiment and the chances of producing positive reviews at least 90\% of the time. 
\begin{table}[tb]
\centering
\caption{\textbf{Evaluation of quality and topic adherence of \llama generations when controlled for specific topics}. 
Topic adherence is evaluated on 300 prompts spanning 6 topics \phitopic~\citep{wettig2025organizewebconstructingdomains}. We report perplexity (\textbf{PPL}), the average \phitopic score (\textbf{Avg \phitopic}), the fraction of examples for which the topic score is greater than 0.8 in 90\% or more of the generations (\textbf{Topic Prob.}), and the expected minimum topic score (\textbf{Exp. Min. Topic}).  
}
\label{tab:experiments-topic-results}
\begin{tabular}{l cccc}
\toprule
\textbf{Method} & \textbf{Topic Prob.} ($\uparrow$)  & \textbf{Exp. Min. Topic} ($\uparrow$)  & \textbf{Avg \phitopic} ($\uparrow$)  & \textbf{PPL} \\
\midrule
\random	    & 86.20 & 83.91 & 91.87	& 6.16 \\
\beamsearch	& 87.47 & 90.35 & 91.63 & $\mathbf{3.78}$ \\
\bestofn	& 95.40 & 95.18 & 97.52 & 8.42 \\
\ours (ours) & \textbf{98.40} & \textbf{96.71} & $\mathbf{99.07}$ & 7.39\\
\bottomrule
\end{tabular}
\end{table}
\subsection{Controlled Topic Generation}
\label{ssec:topic-control-experiments}

Lastly, we evaluate the methods on their ability to control for the topic of LM generations. 
We choose 6 diverse topics from the recently taxonomy concerning the web structure~\citep{wettig2025organizewebconstructingdomains}, including freque (\eg \textit{Finance \& Business} and \textit{Politics}) and less frequent topics (\eg \textit{History}, \textit{Industrial}). 
For each topic, we randomly select 50 different examples from the \topicdataset~\citep{wettig2025organizewebconstructingdomains} test set, breaking them into prefixes of 8 to 12 words. 
Each prefix is used to sample a maximum of 60 tokens.%

\textbf{Topic Generation Task.} 
In general, we find that uncontrolled baselines achieve a fairly high average constraint score ($\geq$ 91\%), which may be explained by the use of longer prefixes during generation. We find this to be the case for most examples (see examples in Appendix \ref{app:additional-results-topic-control}).
Nonetheless, the discrepancy between uncontrolled and controlled methods is still visible with the latter achieving 7\%-8\% higher average constraint scores.
Remarkably, we find \ours is not only able to improve upon \bestofn, achieving an average score of 98.89\% but also produces higher quality generations as emphasized by the lower perplexity.%
\section{Conclusion}
\label{sec:conclusion}
In this paper, we introduced a training-free approach to semantic control of autoregressive
language models.
Our approach uses exact inference on an approximate distribution induced by an LM generation,
using first-order information from a verifier to compute the expected constraint satisfaction
for each of the possible next tokens.
Our approach demonstrated a substantial improvement compared to previous approach on the tasks of controlling the toxicity, sentiment and topic of LM generations.

\section*{Ethics Statement}

Our work investigates the problem of exerting semantic control over LM generations.
While our method can be very societally beneficial, giving us more control over language models, we acknowledge that our method could be misused to produce harmful content.
We look forward to exploring future work that places guardrails on LMs to prevent these
pitfalls.

\section*{Acknowledgments}
This work is supported by the DARPA ANSR
program FA8750-23-2-0004, an NSF CAREER award number IIS-2046873 and an NSF award number 1900644. The conclusions are of the authors and do not reflect the official policy or position of DARPA or the U.S. Government.

\section*{Statement of Author Contributions}
\textbf{Kareem Ahmed}: Conceived and developed the core research idea and the proposed approach. Wrote the introduction and technical sections of the paper. Wrote the code for computing the expected embedding and contributed to debugging the overall approach.

\textbf{Catarina G. Belem}: Implemented the primary codebase. Conducted all experiments and wrote the corresponding experimental section of the paper in addition to the related works.

\textbf{Padhraic Smyth and Sameer Singh}: Senior project leadership. Provided mentorship, supervision, and advisory support throughout the project. Offered critical feedback on the methodology and the manuscript.
All authors read and approved the final manuscript.

\bibliography{colm2025_conference}
\bibliographystyle{colm2025_conference}
\newpage
\appendix

\section{Experiment Details}
\label{appdx:sec:additional-details}

\textbf{\ours. }
As a trade-off between efficiency and performance, we perform exact inference over the top-10 tokens of the base LM. %
For each prefix, we run 2 independent, non-blocking Gibbs Sampling chains for 20 iterations, applying a thinning factor of 5. 
Each chain starts by sampling 25 tokens from the base LM using a combination of nucleus and min-p sampling (\texttt{top\_p}=0.9, \texttt{min\_p}=0.1)~\citep{Holtzman2020The,minh2025turning}. 
A BERT-based model~\citep{modernbert} is used to efficiently approximate the conditionals $\condmarg$.

\section{Hyperparameters Configurations}
\label{appdx:additional-details:methods}

In this section, we describe the hyperparameters used for each of the decoding algorithms and baselines used in this work.
Except where explicitly mentioned we rely on the HuggingFace's implementation\footnote{\url{https://huggingface.co/} (version 4.49.0)} and the default configurations. 

\begin{itemize}
\item \textbf{Random Search} (\random): \texttt{do\_sample=True} 
\item \textbf{Beam Search} (\beamsearch): \texttt{do\_sample=True}, \texttt{num\_beams=5} and \texttt{temperature=0.3}.
\item \textbf{Best-of-N} (BoN) (\bestofn): We implement a custom best-of-n rejection sampling approach~\citep{Stiennon2020-BestOfN-2020-Neurips}, that independently generates $N=10$ sequences using HuggingFace's \texttt{generate} method, parameterized with \texttt{do\_sample=True}, \texttt{top\_p=0.9}, \texttt{min\_p=0.1}. A verifier $\phiattr$ is used to choose the final generation, picking the generation out of the $N$ that maximizes the constraint verifier. 
For the detoxification experiments where the goal is to minimize toxicity as measured by \phitoxicity, we chose the generation that minimizes \phitoxicity (in practice, we maximize $1 - $ \phitoxicity).
\end{itemize}

Experiments were run on RTX A6000 (48GB RAM) GPUs using HuggingFace and PyTorch.

\section{Additional Results}
\label{app:sec:add-results}

\begin{table}[tb]
\small
\centering
\caption{\textbf{Breakdown of the average \phitopic, \textbf{Topic Prob}, and \textbf{Exp.\ Min.\  Topic} for 6 topics when steering \llama generations to adhere to each given topic}. Topics are ordered left-to-right according to their reported frequency in \citet{wettig2025organizewebconstructingdomains}. 
}
\label{tab:avg-phi-attr-by-topic}
\resizebox{\columnwidth}{!}{%
\begin{tabular}{llccccccc}
\toprule
\multicolumn{1}{l}{\multirow{1}{*}{\textbf{Metric}}}&\textbf{Method} & \textbf{Politics} & \textbf{Finance \& Business} & \textbf{Science \& Tech} & \textbf{Food \& Dining} & \textbf{History} & \textbf{Industrial} \\
\midrule
\multirow{4}{*}{\textbf{\phitopic}} &\random & 90.89 & 95.79 & 91.21 & 89.83 & 92.13 & 91.40 \\
&\beamsearch & 90.94 & 97.54 & 86.02 & 90.18 & 91.14 & 93.95 \\
&\bestofn & 97.40 & 98.98 & 98.64 & 94.36 & 98.30 & 97.46 \\
&\ours & \textbf{98.99} & \textbf{99.70} & \textbf{99.42} & \textbf{97.14} & \textbf{99.60} & \textbf{99.56} \\
\midrule
\multirow{4}{*}{\textbf{Topic Prob}}&\random & 84.00 & 92.80 & 84.80 & 84.40 & 84.40 & 86.80 \\
&\beamsearch & 83.60 & 95.60 & 77.60 & 86.80 & 89.20 & 92.00 \\
&\bestofn & 96.00 & 98.00 & 97.20 & 89.60 & 97.20 & 94.40 \\
&\ours & \textbf{98.40} & \textbf{100.00} & \textbf{99.20} & \textbf{93.60} & \textbf{99.60} & \textbf{99.60} \\
\midrule
\multirow{4}{*}{\textbf{Exp.\ Min.\  Topic}}&\random & 82.51 & 92.03 & 81.55 & 82.46 & 82.48 & 82.41 \\
&\beamsearch & 88.51 & 97.08 & 84.61 & 87.64 & 90.36 & 93.89 \\
&\bestofn & 94.93 & 97.74 & 95.58 & 91.52 & 96.21 & 95.13 \\
&\ours & \textbf{96.42} & \textbf{99.08} & \textbf{95.60} & \textbf{93.37} & \textbf{97.47} & \textbf{98.23} \\
\bottomrule
\end{tabular}
}
\end{table}
\subsection{Controlled Topic Generation}
\label{app:additional-results-topic-control}

Lastly, we evaluate the methods on their ability to control for the topic of LM generations. 
We choose 6 diverse topics from the recently taxonomy concerning the web structure~\citep{wettig2025organizewebconstructingdomains}, including frequent (\eg \textit{Finance \& Business} and \textit{Politics}) and less frequent topics (\eg \textit{History}, \textit{Industrial}). 
For each topic, we randomly select 50 different examples from the \topicdataset~\citep{wettig2025organizewebconstructingdomains} test set, breaking them into prefixes of 8 to 12 words. 
Each prefix is used to sample a maximum of 60 tokens.%

\textbf{Topic Generation Task.} 
In general, we find that uncontrolled baselines achieve a fairly high average constraint score ($\geq$ 91\%), which may be explained by the use of longer prefixes during generation. We find this to be the case for most examples.
Nonetheless, the discrepancy between uncontrolled and controlled methods is still visible with the latter achieving 7\%-8\% higher average constraint scores.
Remarkably, we find that \ours is not only able to improve upon \bestofn, achieving an average score of 98.89\% but also produces higher quality generations as emphasized by the lower perplexity.

\section{Efficient Lookahead Generation via Approximate Gibbs Sampling}
\label{sec:Gibbs}
\begin{wrapfigure}{r}{0.5\textwidth}
\hspace{-0.8em}
 \begin{minipage}[t]{\dimexpr.53\textwidth-.5\columnsep}
 \vspace{-2.50em}
    \begin{algorithm}[H]
    \begin{algorithmic}[1]
    \STATE \textbf{Input}: $\texttt{ModernBert}$, prefix $\rvy_{1:i}$, lookahead $\Delta$, block size $B$, num workers $W$, iterations $N$\\\vspace{2pt}
    \STATE \textbf{Output}: $\tilde{\rvy}_{1:T}$ drawn approximately from $p$\\\vspace{2pt}
    \STATE
    \STATE \COMMENT\textcolor{spearmint}{$\triangleright$ Randomly initialize continuation $\rvy_{i+1:T}$}\\
    \STATE $\rvs \gets \texttt{InitializeSequence}(\rvy_{1:i}, \Delta)$\\\vspace{2pt}
    \STATE \COMMENT{\textcolor{spearmint}{$\triangleright$ Launch $W$ workers for $N/W$ updates}}
    \FORALL{\textbf{workers} $w=1$ \textbf{to} $W$ \textbf{in parallel}}
    \FOR{$iter = 1$ \textbf{to} $\lceil N/W \rceil$}\vspace{2pt}
    \STATE \COMMENT{\textcolor{spearmint}{$\triangleright$ Sample block start $j$ in continuation}}\\
    \STATE $j \sim \mathcal{U}(i+1, T-B+1)$
    \STATE $\texttt{blk\_idx} \gets [j:j+B-1]$\\\vspace{2pt}
    \STATE \COMMENT{\textcolor{spearmint}{$\triangleright$ Read (potentially stale) state $\rvs_{local}$}}
    \STATE $\rvs_{\texttt{local}} \gets \texttt{ReadSharedState}(\rvs)$\\\vspace{2pt}
    \STATE \COMMENT{\textcolor{spearmint}{$\triangleright$ Get approximate block conditionals}}
    \STATE $\p_{\texttt{blk}} \gets \texttt{ModernBert}(\rvs_{\texttt{local}}, \texttt{blk\_idx})$\\\vspace{2pt}
    \STATE \COMMENT{\textcolor{spearmint}{$\triangleright$ Sample new tokens for the block}}
    \STATE $\rvy'_{blk} \gets \texttt{SampleFromBlockDist}(\p_{\texttt{blk}})$\\\vspace{2pt}
    \STATE \COMMENT{\textcolor{spearmint}{$\triangleright$ Update shared sequence (Hogwild!)}}
    \STATE $\texttt{WriteSharedState}(\rvs, \texttt{blk\_idx}, \rvy'_{\texttt{blk}})$\\\vspace{2pt}
    \ENDFOR
    \ENDFOR
     \STATE \texttt{WaitForAllWorkers}()
     \STATE $\tilde{\rvy}_{1:T} \gets \texttt{ReadSharedState}(\rvs)$
    \STATE $\mathtt{return}$ $\tilde{\rvy}_{1:T} $\vspace{0.1em}\label{line:return} 
\end{algorithmic}
\caption{Hogwild! Gibbs Sampling}%
\label{algo:gibbs-sampling}
\end{algorithm}
\end{minipage}
\vspace{-1.5em}
\end{wrapfigure}
\label{ssec:gibbs-sampling-efficient-sample-generation}
Our approach requires access to plausible future continuations, or lookahead samples, $\rvy_{i+1:T}$, given a prefix $\rvy_{1:i}$.
However, we would like to avoid expensive autoregressive sampling, especially
since we are happy to trade off sample quality for efficiency.
Intuitively, we are only interested in a crude projection of where the current trajectory might lead us, as opposed to a perfectly coherent natural language sentence.

Taking cue from speculative decoding~\citep{SpeculativeDecoding}, given a prefix $\rvy_{1:i}$
we start with a guess for the continuation $\rvy_{i+1:T}$, either by padding with \texttt{[MASK]} tokens or crudely sampling $\p(\rvy_{j} \mid \rvy_{1:i})$ for $j= i+1$ to $T$.
We can then refine these crude continuations using \emph{Gibbs Sampling}~\citep{koller2009probabilistic}, a Markov chain Monte Carlo (MCMC) approach that stochastically samples each token in the sequence, asymptotically converging to the true
distribution.
Therefore, by setting a \emph{cutoff}, or a maximum number of iterations, we can control
how crude of a lookahead sample we desire.
Unfortunately, this introduces a multitude of computational challenges.
First, the Gibbs sampler assumes efficient access the the full conditionals $\p(\rvy_{i} \mid \rvy{-i}) \forall i$, which requires $O(|\sV|)$ forward passes of the LM for a single
position $i$, which is untenable given the vocabulary size of modern LMs.
Second, in its most basic form, Gibbs sampling requires many iterations through the sentence, computing the conditional and resampling a single token per iteration, which is quite slow.

To overcome these challenges and enable efficient generation, we utilize several strategies:

\textbf{Approximate Conditionals with Masked Language Models (MLMs)\ } In place of analytically computing the conditionals computation, we leverage efficient pretrained MLMs to approximate the conditional probability $p(\ry_i|\rvy_{-i})$.
These models are inherently designed to predict masked tokens given their bidirectional context, providing a fast approximation of the required conditional distributions without expensive analytical marginalization.

\textbf{Parallel and Asynchronous Updates (Hogwild! Style)\ } Standard Gibbs sampling updates tokens sequentially.
In a bid to accelerate sampling, we employ parallel, potentially asynchronous updates inspired by Hogwild!~\citep{HogwildGibbs, HogwildSGD} approaches. 
Multiple token positions $j$ can be updated simultaneously, possibly using slightly stale context information $\rvy_{-j}$.
This trades off the unbiasedness of Gibbs sampling~\citep{HogwildBias} for substantial gains in wall-clock time tht are crucial for inference-time applications.

\textbf{Blocked Gibbs Sampling\ } 
Rather than sampling individual tokens one at a time, we can update contiguous blocks of tokens simultaneously.
This reduces the number of sampling iterations required for convergence of the chain while allowing us to better leverage the parallel processing capabilities of modern hardware, especially when combined with MLM-based approximate conditionals that excel at processing multiple positions efficiently.

\textbf{Controlling the Efficiency-Accuracy Trade-off\ } The use of approximate conditionals introduces a natural dial to balance efficiency and sample quality.
In very much a Hogwild! fashion, the frequency at which we re-compute or synchronize these approximate conditionals using the latest context influences this trade-off. Less frequent updates lead to faster sampling using potentially more outdated contextual information, while more frequent updates improve fidelity to the target distribution at the cost of increased computation.

By combining Gibbs sampling with these efficiency-focused techniques—approximating conditionals via MLMs, parallelizing updates Hogwild! style, and employing blocked sampling—we can rapidly generate diverse and plausible lookahead samples $\rvy_{i+1:T}$ suitable for our inference-time algorithm, effectively transforming the computationally demanding task of sampling from the joint distribution into a manageable and efficient procedure.

The pseudocode for the approach elucidated above can be seen in~\cref{algo:gibbs-sampling}. Furthermore, an efficient PyTorch implementation will be made available in our GitHub Repository.

\end{document}